\documentclass[acmsmall,screen,authorversion]{acmart}

\newcommand{\edit}[1]{\textcolor{black}{#1}}
\newcommand{\stat}[1]{\textcolor{black}{#1}}
\newcommand{\fredit}[1]{\textcolor{black}{#1}}

\usepackage{multirow}
\usepackage{booktabs}
\usepackage{stfloats}

\AtBeginDocument{%
}

\newcommand{\VanderbiltAffil}{%
\affiliation{%
\institution{Vanderbilt University}
\city{Nashville}
\state{TN}
\country{United States}}}

\newcommand{\VUMCAffil}{%
\affiliation{%
\institution{Vanderbilt University Medical Center}
\city{Nashville}
\state{TN}
\country{United States}}}

\newcommand{\ClemsonAffil}{%
\affiliation{%
\institution{Clemson University}
\city{Clemson}
\state{SC}
\country{United States}}}

\newcommand{\UTHealthHoustonAffil}{%
\affiliation{%
\institution{University of Texas Health Science Center at Houston}
\city{Houston}
\state{TX}
\country{United States}}}

\newcommand{\UICAffil}{%
\affiliation{%
\institution{University of Illinois Chicago}
\city{Chicago}
\state{IL}
\country{United States}}}

\newcommand{\ICSIAffil}{%
\affiliation{%
\institution{International Computer Science Institute}
\city{Berkeley}
\state{CA}
\country{United States}}}

\title[Advice Differences in Alzheimer's Caregiving Support]{One Question, Four Voices: How Advice for Alzheimer's Caregiving Differs Between Caregivers, Clinicians, and Large Language Models}

\author{Congning Ni}
\email{congning.ni@vanderbilt.edu}
\orcid{0000-0001-6950-6948}
\VanderbiltAffil
\VUMCAffil

\author{Jinkyung Katie Park}
\email{jinkyup@clemson.edu}
\orcid{0000-0002-0804-832X}
\ClemsonAffil

\author{Yang Li}
\email{yang.li.2@vanderbilt.edu}
\orcid{0009-0003-5177-1796}
\VanderbiltAffil

\author{Sarvech Qadir}
\email{sarvech.qadir@vanderbilt.edu}
\orcid{0009-0006-7962-5792}
\VanderbiltAffil

\author{Sean S. Huang}
\email{sean.huang@vumc.org}
\orcid{0000-0002-5050-2012}
\VUMCAffil

\author{James S. Powers}
\email{james.powers@vumc.org}
\orcid{0000-0002-5305-602X}
\VUMCAffil

\author{Julia A. Hiner}
\email{julia.a.hiner@uth.tmc.edu}
\orcid{0009-0003-5226-2141}
\UTHealthHoustonAffil

\author{Vania Leung}
\email{vleung3@uic.edu}
\orcid{0009-0001-4734-3639}
\UICAffil

\author{Patricia Commiskey}
\email{patricia.commiskey@gmail.com}
\orcid{0000-0001-6376-4698}
\VUMCAffil

\author{Lijun Song}
\email{lijun.song@vanderbilt.edu}
\orcid{0000-0003-3162-2649}
\VanderbiltAffil

\author{Qingxia Chen}
\email{cindy.chen@vumc.org}
\orcid{0000-0003-0507-1847}
\VUMCAffil

\author{Pamela J. Wisniewski}
\email{pwisniewski@icsi.berkeley.edu}
\orcid{0000-0002-6223-1029}
\ICSIAffil

\author{Bradley A. Malin}
\email{b.malin@vumc.org}
\orcid{0000-0003-3040-5175}
\VUMCAffil
\VanderbiltAffil

\author{Zhijun Yin}
\email{zhijun.yin@vumc.org}
\orcid{0000-0002-3075-1337}
\VUMCAffil
\VanderbiltAffil

\renewcommand{\shortauthors}{Ni et al.}

\received{May 13, 2025}
\received[revised]{January 13, 2026}
\received[accepted]{April 9, 2026}

\begin{document}

\setcopyright{cc}
\setcctype{by}
\acmJournal{PACMHCI}
\acmYear{2026} \acmVolume{10} \acmNumber{6} \acmArticle{CSCW132}
\acmMonth{10} \acmDOI{10.1145/3816980}

\begin{abstract}
Online caregiver forums are a \edit{critical support infrastructure} for Alzheimer’s disease and related dementias (ADRD), where people seek both emotional reassurance and practical guidance. \edit{Prior CSCW and HCI work has examined peer and clinician support in these communities and, separately, explored support-oriented chatbots and AI-assisted helping; however, we still lack systematic evidence on how AI responses compare with peer and clinician roles when answering the same caregiver questions in situ.} We address this gap through a systematic, side-by-side comparison of responses to 85 real caregiver questions from \textit{ALZConnected} across four responder groups: \textit{peer caregivers}, \textit{physicians}, \textit{ChatGPT}, and a retrieval-augmented \textit{CareGPT} grounded in peer discussions. We evaluate responses using 28 measures spanning linguistic form (psycholinguistic cues, structure, and style), emotional support, informational support, and topic scope, \edit{including LLM-based scoring for selected measures that we audit and calibrate against double-coded human ratings}. Across analyses, we observe a consistent division of communicative labor: peers provided time-anchored narratives and community grounding; clinicians provided concise, advice-forward, clinically bounded guidance; and LLMs produced a formal, verbose, highly structured ``consultation'' voice. Notably, CareGPT shows minimal measurable separation from ChatGPT, indicating that retrieval-plus-prompt-injection alone is insufficient to shift the model’s communicative stance. Quantitatively, LLM outputs score higher than peers on politeness and emotion-support markers, and are comparable to clinicians on informativeness and physician-alignment measures; yet, they are less narrative and less contextually situated. Topic modeling further shows that LLM responses concentrate narrowly on advice-oriented themes, while peers cover a broader range (e.g., emotional processing, legal/financial concerns, and daily cares). These insights show that AI systems can complement, but not replace, human contributions to ADRD caregiver support. We advocate for hybrid systems that combine AI's structured delivery with \fredit{the distinct contributions of} human voices to strengthen emotional and informational caregiving support infrastructures, while future work should examine whether similar patterns hold in other caregiving domains.
\end{abstract}




 \maketitle

\section{Introduction}

Alzheimer’s caregiving is a profoundly demanding endeavor \edit{that requires a tremendous amount of care coordination}, often undertaken by informal caregivers such as family members of those living with Alzheimer’s disease or related dementias (ADRD). In the United States alone, an estimated 11 million informal caregivers provide unpaid care to someone with ADRD at home \cite{hasan2024empowering}. Caregivers routinely face emotional stress, logistical complexity, and informational uncertainty \cite{smriti2024emotion}, prompting many to seek connection and guidance through online peer support communities. \edit{CSCW researchers have articulated that such online communities operate as critical \emph{care infrastructure} \cite{karusala2023unsettling}, sustaining day-to-day coordination, sense-making, and mutual support when formal resources are limited \cite{allison2021logging, johnson2022s, poon2023computer}. We contextualize our research in such a community, one called \textit{ALZConnected}, where ADRD caregivers exchange advice, share lived experiences, and offer emotional validation to one another. Consistent with social support theory \cite{scott2014social}, these interactions often combine emotional support (e.g., empathy and solidarity) \cite{gencc2024situating} with informational support (e.g., experiential knowledge and practical suggestions) \cite{ni2025catalysts}. Increasingly, such caregiving support infrastructures are also becoming sites where emerging AI systems participate alongside peers and professionals, reshaping how guidance and care are produced and encountered \cite{park2025current, wang2021cass, meng2023mediated, wester2024chatbot}}

%However, peer-generated responses can vary in clinical precision and consistency \cite{mendel2024advice}, and many caregivers are themselves overwhelmed, leading to unanswered or delayed posts. 
\edit{Growing interest in AI-generated support as a scalable, on-demand supplement within caregiving communities stems from structural limitations in peer support, as peer-generated responses can vary in clinical precision and consistency \cite{mendel2024advice}, and many caregivers are themselves overwhelmed, leading to unanswered or delayed posts \cite{yin2024perceptions}.} In contrast, professional responders, such as physicians, are trusted sources of medical expertise, but their limited availability and formal communication norms often constrain their presence in everyday online caregiving spaces and often limit deeper engagement even within clinical encounters\cite{ding2020getting}. \edit{From a CSCW perspective, these human and institutional limitations lead to the need for \emph{care infrastructures} in which support work is distributed across family members, peer communities, and professionals \cite{chen2013caring,kabir2025balancing}.} Emerging LLM-based AI systems (e.g., ChatGPT) now add an easy-to-access source of on-demand guidance and emotionally supportive replies in health-related contexts, yet prior evidence highlights inconsistent reliability and calls for the need for careful evaluation of AI-mediated healthcare conversations \cite{wang2024applications,chin2023potential,abbasian2024foundation}. %The urgent CSCW challenge is not simply whether AI can produce fluent replies, but \textbf{how caregivers \emph{calibrate reliance} \cite{mcdonald2019reliability} across peers, clinicians, and AI when emotional support and practical health guidance come together in the same interaction}. 
\edit{This shift raises a core CSCW issue in hybrid care infrastructures: Caregivers increasingly receive support from peers, clinicians, and AI systems whose responses may differ in language, emotional framing, and informational content. Understanding these communicative differences is essential for designing socio-technical systems that deliver timely, relevant, and efficient support for everyday caregiving work \cite{zubatiy2023don,khullar2025nurturing,chen2013caring,kabir2025balancing,petsolari2024sociotechnical}.}

% how support roles are divided, when AI-generated guidance fits with or conflicts with human contributions, and what interaction pathways help caregivers check information and decide what to do under uncertainty 

Prior work has begun to examine how different support sources compare in the kinds of help they provide, but evidence remains fragmented across separate lines of work. \edit{Studies of online health communities characterize peer responses as rich in lived experience, storytelling, and emotional affirmation, yet also uneven in timeliness and informational rigor \cite{daynes2023online,ni2023examining}.} Complementary work on clinician involvement highlights how professional expertise can improve clinical precision, while also showing that participation is constrained by limited availability and clinical communication norms \cite{li2025bridging}. \edit{More recently, evaluations of LLM-mediated health support compare model-generated replies with human guidance and document a trade-off between scalable, well-structured responses and risks of factual errors, missing context, or unclear provenance \cite{huang2024fact,ayers2023comparing}. Building on these themes, recent CSCW/HCI work further motivates the need to scrutinize how conversational agents function within caregiving networks and how evaluation practices should reflect real-world support needs \cite{zubatiy2023don,khullar2025nurturing}.} At the same time, it is increasingly common to augment general-purpose LLMs with retrieval-augmented generation (RAG) to ground responses in domain-relevant evidence, motivating systems such as our designed \textit{CareGPT}. However, we still lack a \emph{side-by-side} characterization of how peers, clinicians, and LLM-based systems (with or without RAG) differ when responding to the \emph{same} caregiver needs, and \edit{how such differences should inform the design of hybrid care infrastructures \cite{karusala2023unsettling} in which multiple support voices co-exist.}

To fill this gap, we conduct a systematic comparison of how four support categories respond to real caregiver questions posted on \textit{ALZConnected}: (1) online peer caregivers, (2) board-certified physicians, (3) ChatGPT, and (4) CareGPT, implemented by enhancing GPT-4 with RAG over online caregiving discussions. We select these four support categories to reflect a hybrid \emph{care infrastructure} in which support is produced through multiple roles and tools: peer experience, clinical expertise, and AI systems that are increasingly available alongside them \cite{chen2013caring,kabir2025balancing}. Rather than ranking responders, we examine how AI responses align with or diverge from human communication when caregivers may encounter different sources for the \emph{same} high-stakes question. \edit{Guided by care infrastructures \cite{karusala2023unsettling}}, we examine how different responders contribute to caregiving communication using three complementary dimensions: linguistic characteristics, social support functions, and topical focus. \edit{In our evaluation, we chose support dimensions that aligned with social support theory, motivating our focus on \emph{emotional} and \emph{informational} support as core support functions in online scenarios \cite{scott2014social}. We include linguistic features to capture how responses are delivered (e.g., structure, readability, affective framing) and topic patterns to characterize how caregiving needs are addressed across responders \cite{mamykina2015collective}.} Therefore, we organize our research around three research questions:

\begin{itemize}
    \item \textbf{RQ1:} How do responses from peer caregivers, physicians, and LLM-based systems (ChatGPT, CareGPT) differ in linguistic characteristics \edit{(e.g., structure, style, and psycholinguistic cues)?}
    \item \textbf{RQ2:} How do these responder groups differ in emotional and informational support \edit{(e.g., reassurance/validation and advice/resources)} when responding to caregiver questions?
    \item \textbf{RQ3:} \textit{How do \edit{linguistic and support-related patterns} vary across \textbf{caregiving topics}, and how do these variations differ by responder group?}
\end{itemize}

To investigate these questions, we analyze 85 caregiver-posted questions from \textit{ALZConnected}\footnote{https://alzconnected.org/discussions/tagged/Asking-for-Help}, each paired with four responses, with one per responder group. This question-controlled design enables systematic comparison of how peers, clinicians, and LLM-based systems respond to the same caregiving needs, treating each response as a situated act of care within a broader caregiving infrastructure. We operationalize linguistic structure (e.g., verbosity, readability), psycholinguistic markers (e.g., emotional tone, cognition \& perception), emotional and informational support cues (e.g., empathy, information evaluation), and topical focus (via BERTopic modeling), as observable traces of how care is enacted, coordinated, and made legible across different support roles. We develop NLP tools and LLM-based scoring pipelines that we validate through human evaluation metrics to ensure reliability and contextual sensitivity, aligning quantitative measures with CSCW theories of care, support work, and sociotechnical participation. Our contributions to CSCW and human-AI interaction research are \edit{three-fold}:

\begin{itemize}
    \item We empirically show that peer caregivers, clinicians, and LLM-based systems enact caregiving support through systematically different communicative roles, \fredit{revealing that caregiving support is organized through distinct but complementary forms of relational, informational, and accountable labor within a hybrid care infrastructure, rather than convergence toward a single ``best'' form of support.}
    
    \item We demonstrate how emotional and informational support are differently balanced across responder groups, with peers grounding care in lived, narrative experience, clinicians providing bounded and clinically oriented guidance, and AI systems offering structurally consistent, polite, and advice-forward responses. \fredit{This extends social support theory by showing how AI operates as a distinct infrastructural participant within caregiving support ecosystems.}
    
    \item \edit{We show that these differences are not uniform, but vary meaningfully across caregiving topics, indicating that caregiving support is situationally enacted and that the value and limitations of AI, peer, and clinician responses depend on the nature of the caregiving concern.}

\end{itemize}

Overall, these findings \fredit{advance CSCW discussions of care infrastructures by characterizing how support work is redistributed across human and AI actors through distinct relational roles and trust cues}, and inform the design of hybrid caregiving systems that coordinate complementary roles rather than treating AI as a replacement for human care in online ADRD caregiving support communities.

While we focus on ADRD caregiving, \edit{this infrastructure-oriented approach} can be extended to other caregiving and health support settings where people increasingly navigate hybrid human-AI support ecosystems.

\section{Related Work}
\subsection{\edit{Care Infrastructures and Support Functions in CSCW Research}}
\label{sec:theory}
\edit{CSCW research has long conceptualized care and support not as isolated, one-off interactions, but as ongoing cooperative work sustained through coordination, participation, and day-to-day maintenance across people and technologies \cite{chen2013caring,karusala2021future}.} Building on this tradition, \edit{recent CSCW scholarship has articulated an \emph{infrastructural care} lens}, which emphasizes how caregiving practices are assembled and strained through heterogeneous support sources such as family members, peer caregivers, clinicians, and sociotechnical systems that collectively enable care under conditions of uncertainty and constraint \cite{dye2021ungrano,bhat2021telehealth,karusala2021future,wilcox2023infrastructuring}. 
In long-term Alzheimer's caregiving contexts, where caregiving needs evolve over time and caregivers often live under high workload, engaging with and interpreting support from multiple sources becomes part of the work of sustaining care itself \cite{massimi2014life,wang2012stay,allison2021logging}. From this perspective, \edit{a key challenge for designing effective hybrid care infrastructures lies not only in the availability of support, but in understanding how different support sources communicate and frame support or assistance across caregiving challenges}. Without a clear empirical understanding of how these responder groups differ in their communicative patterns, it remains difficult to design socio-technical systems that efficiently and appropriately support caregiving needs.

To describe \emph{what kinds} of support are exchanged within these infrastructures, communication research distinguishes multiple support types. 
In text-based online forums, \edit{emotional and informational support are among the most visible and widely investigated forms \cite{langford1997social}. 
CSCW and health communication research further suggest that informational support is closely tied to lay sensemaking and negotiation over what counts as actionable knowledge \cite{huh2014weaving}. }
In ADRD caregiving, support needs often combine practical guidance with emotional strain and self-care tensions, reinforcing why emotional and informational support are central analytical dimensions \cite{shi2025balancing}. 
At the same time, when support is mediated by technology, interactional cues (e.g., assertiveness, hedging, or adaptation to expertise) can shape how guidance is interpreted; similar HCI work in clinician-AI interaction shows communication style should be treated as part of system behavior rather than merely ``packaging'' \cite{calisto2023assertiveness,calisto2025personalized,calisto2024human}. 
\edit{In this paper, we use care infrastructure \cite{karusala2023unsettling} to motivate why it is meaningful to compare peers, clinicians, and AI \emph{together} as co-present sources of support.} We further draw on social support theory \cite{scott2014social} to focus our comparison on emotional and informational support, and examine these dimensions through linguistic characteristics and topic patterns in responses.

\subsection{\edit{Support from Online Peer Caregivers and Clinicians}}
A large body of CSCW and health-community research shows that caregivers rely on online forums to share experiences, seek advice, and cope with long-term caregiving demands \cite{massimi2014life, johnson2020roles, lazar2019safe}. These communities commonly provide emotional support through encouragement and validation \cite{han2022impact}, and informational support through practical strategies and shared resources \cite{wallace2021caregivers}. People living with ADRD often get worsen conditions by gradually losing memory and cognitive ability \cite{cole2023care}, thus demanding far more care than other older adults. Therefore, caregiving discussions often reflect prolonged emotional strain, uncertainty about progression, and ongoing logistical coordination \cite{smriti2024emotion}. Qualitative studies further document recurring tensions around ethics, coping, and family conflict in caregiver discourse \cite{ullrich2020ethical,halevi2022emotion}, including work that highlights how ``informal'' caregiving conversations surface complex responsibilities and strain \cite{foster2025informal}. Large-scale analyses also suggest measurable benefits of peer interaction, such as improved affect following peer responses \cite{ni2024sentiment}. At the same time, caregivers often seek support because they are under prolonged stress and isolation \cite{kang2014factors}, which includes explicit requests for comfort and validation \cite{ni2022rough}.

\edit{Online caregiving infrastructures face informational challenges. Caregivers often need reliable, context-relevant information when decisions can shape care trajectories \cite{mason2022health},} and studies document how caregivers search for and evaluate information under uncertainty \cite{scharett2017investigation}. Clinicians are authoritative sources, but their responses can be brief or sometimes difficult to interpret for lay audiences \cite{hamilton2006physicians}, and sustained participation in everyday online support settings is often constrained by time, institutional boundaries, and professional communication norms \cite{ding2020getting}. A study on clinician participation shows how professional input can complement peer experience, while still being difficult to sustain in practice \cite{li2025bridging}. Peer caregivers provide experiential knowledge that can feel more relatable to caregiving experience, but advice quality can vary in evidence and generalizability \cite{brock2017peer}, including uneven clinical precision and consistency \cite{mendel2024advice}.

Care support, however, is not only about exchanging facts and knowledge; it also depends on how people respond to each other. \edit{CSCW research emphasizes the affective component of caregiving communication in chronic illness contexts \cite{bhat2023we}, including conversational practices that accomplish empathy through concrete interactional moves \cite{pudlinski2005doing}.} More broadly, scholarship on socio-emotional labor describes the effort required to manage one’s own emotions while responding with care \cite{leana2018more}. In online communities, this can appear through storytelling and responsive mirroring \cite{larson2005clinical}, and repeated peer support practices that sustain participation over time \cite{poon2021computer}. Such support can matter for relationship quality and outcomes for support seekers \cite{verhofstadt2016role}. Conversely, messages that feel blunt or automated can reduce perceived warmth and credibility \cite{fauchon2017does}. Similarly, while clinicians provide valued professional guidance, their expressions of empathy are often framed through clinical expertise rather than shared lived experience, which can differ from the emotionally grounded support offered by peer caregivers. These studies motivate a question-controlled comparison of how different support categories respond to the same caregiving needs in terms of linguistic delivery, topical patterns, and support types \cite{shi2025balancing}.

\subsection{\edit{LLMs in Caregiving Support}}
\label{sec:AIsupport}
Recent advances in LLMs have renewed interest in AI as a participant in health-related communication. In some settings, evaluators rate LLM answers as helpful and empathetic, and may even prefer them over clinician-written answers on perceived quality \cite{ayers2023comparing}. Related work finds that LLMs can produce emotion-aware and socially appropriate responses across different health question formats \cite{sorin2024large,zhao2023chatgpt}, including evaluations of ChatGPT-style models for assessment and response generation in health contexts \cite{yeo2023assessing}. Researchers have also explored LLM-based systems that aim to better model supportive dialogue, including empathy-oriented tuning \cite{chen2023soulchat}, alongside broader evaluations of LLM capabilities for health communication \cite{chen2023llm}. In caregiving-adjacent scenarios, interaction style also matters: caregivers and lay users may read coherence, politeness, and clarity as cues of support \cite{zou2024pilot}.

At the same time, research cautions that AI support can feel generic, overly polished, or simply mismatched to sensitive situations \cite{connolly2012cscw, oppermann2019beyond}. Studies of AI-mediated emotional support raise concerns about authenticity and ``performed'' empathy \cite{rubin2024considering}, including analyses of how systems are designed to \emph{create} or signal empathy in interaction \cite{schultz2023creating}. Related findings suggest that users may accept chatbot support for everyday stress while still preferring human support in harder moments \cite{young2024role}. In caregiving networks, limitations often surface when nuanced, context-sensitive support is most needed \cite{zubatiy2023don}. This has motivated hybrid designs that try to \emph{augment} human supporters rather than replace them: for example, Sharma et al.\ introduce a system that gives just-in-time feedback to peer supporters to help them craft better replies while keeping humans in the loop \cite{sharma2023human}. \edit{More broadly, HCI studies show that communication cues are part of system behavior, which is work on agent assertiveness \cite{calisto2023assertiveness}, explanations adapted to expertise levels \cite{calisto2025personalized}, and human-centered guidance for personalized agents \cite{calisto2024human} all point to how framing, tone, and explanation choices can shape user judgment and downstream action.}

A further concern is reliability in high-stakes advice settings. LLMs can generate confident but incorrect statements \cite{aljamaan2024reference}, which has prompted calls to prioritize safety and trustworthy evaluation over fluency \cite{au2023ai}. Related work discusses practical evaluation concerns and failure modes for AI-mediated healthcare conversations \cite{abbasian2024foundation}, as well as tensions around trustworthiness and normative fit when AI enters support communities \cite{chin2023potential}. Researchers have explored RAG and other verification-oriented mechanisms to improve grounding and transparency \cite{ferrario2022explainability}. Work on how people judge AI outputs emphasizes interpretability and perceived reliability cues \cite{johnson2023assessing}, and studies of algorithm-mediated services show that users develop ``folk theories'' and strategies under uncertainty \cite{lopez2025users}. \edit{Finally, adjacent HCI work on emotion-focused technologies cautions against overclaiming what AI can do in affective domains \cite{liu2022will}, and argues for treating emotion regulation and care as situated practices rather than generic text outputs \cite{liu2025regulation}. Together, these concerns motivate evaluating LLM responses as one voice in a broader care infrastructure, not as a standalone replacement to human support providers. From a CSCW perspective, this framing foregrounds caregiving support as ongoing, distributed work accomplished through heterogeneous actors and technologies. Building on this view, our study examines how AI systems, peer caregivers, and clinicians differentially enact care within a shared support infrastructure when responding to the same caregiver requests. We next describe our study design and evaluative approach that operationalizes and compares this infrastructural caregiving perspective.}

\section{Methodology}
\edit{Guided by \emph{care infrastructure} (\S\ref{sec:theory})}, we designed a question-controlled comparative study to examine how different responder groups provide support within an online caregiving setting. We compare peer caregivers, clinicians, and LLM-based systems, evaluating responses along linguistic characteristics, emotional and informational support, and topic-specific patterns.

\begin{figure}[ht]
    \centering
    \includegraphics[width=0.9\linewidth]{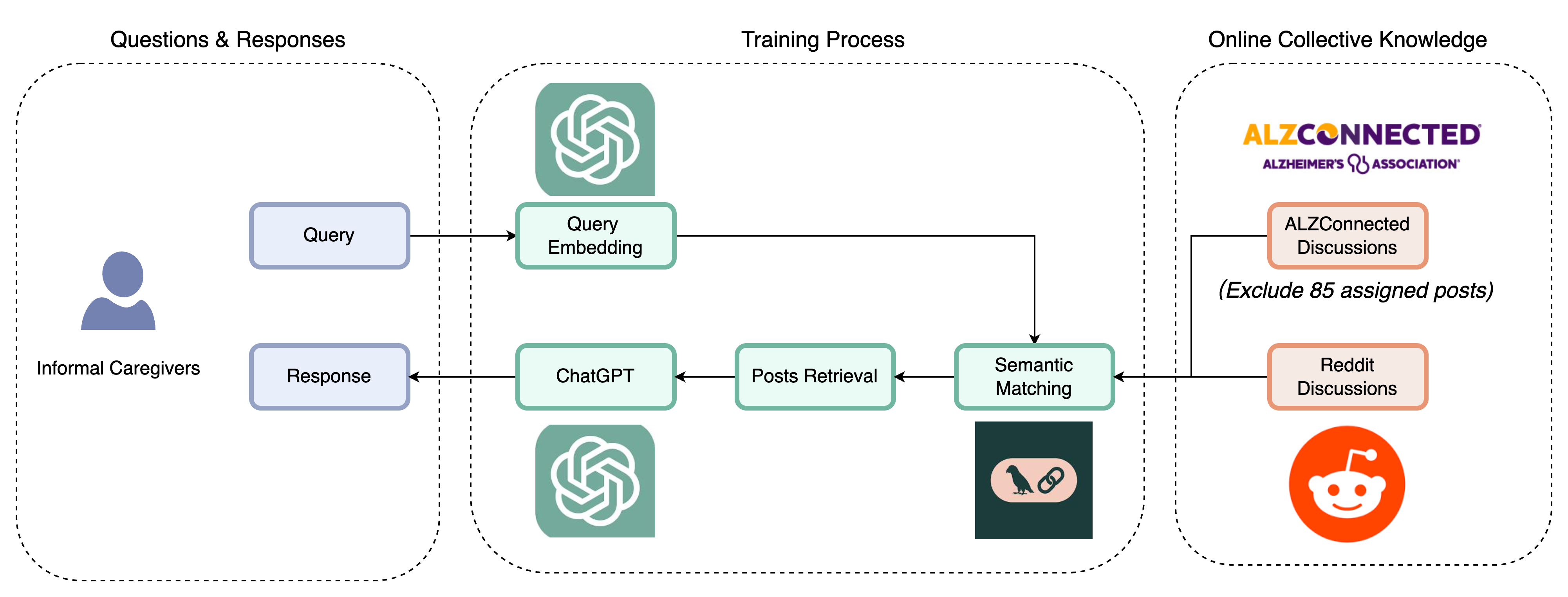}
    \caption{
    The design of CareGPT, including query embedding, semantic matching, relevant post retrieval, and response generation.
    }
    \Description[Pipeline diagram showing CareGPT’s system workflow]{
    A user submits a free-text question, which is embedded via an OpenAI embedding model. This vectorized query is used to retrieve semantically similar posts from two sources, ALZConnected and Reddit caregiver forums. In parallel, ChatGPT generates a standalone response. Retrieved posts and the ChatGPT output are passed through a semantic matcher to identify the most contextually relevant content for final delivery.
    }
    \label{fig:pipeline}
\end{figure}

\subsection{Implementation of CareGPT}
\label{subsec:caregpt-impl}
We developed CareGPT as a LangChain-based RAG system tailored to dementia caregiving. Fig.~\ref{fig:pipeline} illustrates the pipeline of CareGPT. Unlike vanilla ChatGPT, CareGPT grounds its responses in semantically retrieved content from real online peer discussions, enabling the model to draw on community-situated experiential knowledge that circulates within caregiving forums.

The system indexes a curated knowledge base constructed from 50{,}000+ posts (\textbf{excluding} the 85 authentic caregiver questions for our study) and 300{,}000+ comments drawn from ALZConnected and Reddit caregiving forums. These discussions were embedded using OpenAI's \texttt{text-embedding-ada-002-v2} model and stored in a FAISS vector database for semantic search. A caregiver’s query is embedded, and the top-$k$ ($k = 4$) semantically similar posts, together with their associated comments, are retrieved. The retrieved content is anonymized using \texttt{presidio} \cite{mendels2018microsoft} before being injected into GPT-4 via structured prompts. To minimize topical drift from long documents, posts are embedded independently rather than concatenated into chunks, thereby preventing unrelated comments from being conflated during retrieval.

Since the quality of RAG systems depends on both retrieval and generation, we manually assessed the retrieval performance. To ensure broad topical coverage, we sampled a stratified subset of caregiver questions. Two annotators independently labeled each retrieved post along two binary dimensions: (1) whether the post was topically \textbf{relevant} to the caregiver’s query, and (2) whether it showed observable \textbf{influence} on the generated CareGPT response. Annotator agreement was then summarized using percent agreement and Cohen’s $\kappa$.

\subsection{Study Design and Dataset}
\label{subsec:study-design}

\edit{We conducted a mixed-methods, question-controlled comparative study to examine how different support categories respond to the \emph{same} caregiver needs (i.e., based on requests posted on the \textit{ALZConnected} forum) within an online caregiving infrastructure}. Each caregiver question was paired with one response from four distinct responder groups, allowing us to compare how support is delivered across linguistic characteristics, emotional and informational support functions, and topic-specific patterns.

We curated 85 anonymized caregiver questions from the ``\textit{Asking for Help}'' board of the \textit{ALZConnected} forum\footnote{https://alzconnected.org/discussions}, posted between April 7 and October 16, 2023. These posts are publicly accessible, and for this study we collected only textual content and removed usernames and other obvious identifiers prior to analysis. Within this time window, we included \textit{all} advice-seeking posts that (a) were written in English by informal caregivers, (b) were first-person requests for help, and (c) had at least one peer reply, yielding 85 eligible questions. This final 85 questions, therefore, reflects the complete set of posts that met our inclusion criteria while remaining feasible for collecting matched physician and AI responses for each question.

Each question was matched with four responses from the following responder groups:
\begin{itemize}
    \item \textit{Online peer caregivers}, authentic replies posted by community members on \textit{ALZConnected};
    \item \textit{Physicians}, answers authored by four board-certified physicians recruited to contribute expert responses for this study;
    \item \textit{ChatGPT-4}, responses generated by OpenAI's GPT-4 model;
    \item \textit{CareGPT}, responses from the GPT LangChain-based RAG system as described in \S\ref{subsec:caregpt-impl}.
\end{itemize}

We selected GPT-4 due to its prominence and extensive evaluation in health and well-being research at the time of this study \cite{wang2024applications}. To keep communicative intent comparable across non-peer responders, we used a single standardized instruction for both physicians and AI models: \textit{``Envision yourself in your office with the caregiver and provide an answer that is both empathetic and accurate.''} \edit{This prompt establishes a shared consultation scenario without prescribing specific wording; we interpret stylistic differences (e.g., degree of formality) as properties of this controlled setup rather than fixed traits of any responder in all contexts. Peer caregiver responses were drawn directly from the forum and anonymized prior to analysis.}

\begin{figure}[ht]
    \centering
    \includegraphics[width=0.6\linewidth]{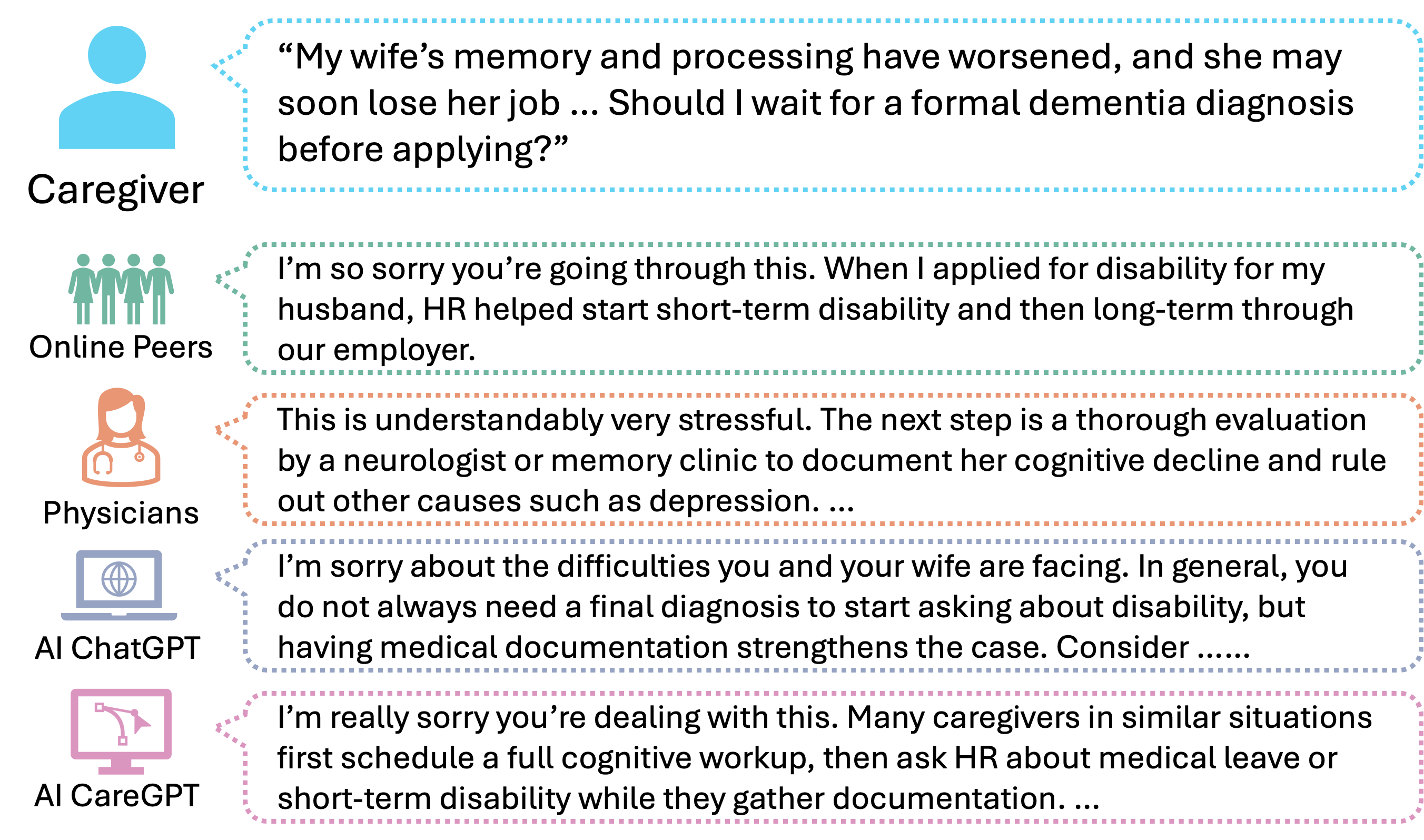}
    \caption{Example caregiver question (top) and corresponding responses from online peer caregivers, a physician, ChatGPT-4, and CareGPT (bottom). The figure mirrors the conversational structure of our dataset, in which one caregiver question is paired with four responder groups. Text has been lightly paraphrased and truncated for readability and to preserve anonymity.}
    \Description[Example caregiver question and four responder-group replies]{
    A schematic conversation example showing one anonymized caregiver question at the top and four corresponding response types below: online peer caregivers, a physician, ChatGPT-4, and CareGPT. The figure illustrates the study design in which the same caregiver question is paired with responses from four responder groups.
    }
    \label{fig:example-conversation}
\end{figure}

Because caregiver threads sometimes contained multiple comments, we treated the set of replies to a given question as the peer response for that item by first computing each metric at the individual-comment level and then averaging these values across comments, yielding one peer-level score per question. For physicians, ChatGPT, and CareGPT, each question received exactly one response (one physician answer and one first-pass generation from each AI model), so metrics were computed directly on these single responses. Figure~\ref{fig:example-conversation} shows one example caregiver question and the corresponding responses from the four responder groups.

The study protocol was reviewed and approved by our Institutional Review Board (IRB). \fredit{Four board-certified physicians with clinical experience in geriatric medicine, older adult care, and dementia-related care were recruited to contribute physician responses to the 85 caregiver questions under the approved study protocol.} We recognize that caregiving discussions involve sensitive emotional content and that community members may not anticipate research reuse in all contexts. To reduce risk, we (1) removed all personally identifiable information (e.g., usernames, contact details), (2) analyzed data only in aggregate and report results at the level of responder groups rather than individuals, (3) paraphrased and truncated any illustrative excerpts to avoid search-based re-identification, and (4) do not release raw post-level text in any public artifact. All data were stored and analyzed on secure institutional infrastructure.

We evaluated each response using 28 predefined features spanning five dimensions: linguistic structure, psycholinguistic markers, emotional support, informational content, and topic relevance. Quantitative comparisons using Analysis of Variance (ANOVA) were conducted across responder groups. To control for Type I error from multiple comparisons, we applied a Bonferroni correction with a conservative threshold of $\alpha = 0.05/28 = 0.0018$, treating results with $p < 0.001$ as statistically significant. \edit{A post hoc power analysis using G*Power \cite{faul2007g} confirmed that our dataset of 340 responses achieved a statistical power of $0.981$ for detecting medium-sized effects in one-way ANOVA, indicating high sufficiency of the response volume for our planned comparisons.}

\subsection{Evaluation Metrics}
\label{sec:metrics}
To support a question-controlled comparison across responder groups, we operationalize each response using a set of measurable communication properties aligned with our three research questions: linguistic characteristics (RQ1), emotional and informational support (RQ2), and topic-specific patterns (RQ3). We organize these measures into six categories: \textbf{Psycholinguistic Characteristic}, \textbf{Linguistic Structure}, \textbf{Linguistic Style}, \textbf{Emotional Support}, \textbf{Informational Support}, and \textbf{Topic Modeling}. Our psycholinguistic and linguistic measures build on foundational work in computational psycholinguistics and language use \cite{tausczik2010psychological,pennebaker2015development} and on text comprehension and readability research \cite{mcnamara2010linguistic}. For emotional and informational support, we draw on social support theory and related operationalizations in online health communication, where these two support types are widely studied and measurable in text \cite{scott2014social,biyani2014identifying,deetjen2016informational,ni2024sentiment}.

The final set of 28 features was chosen from a larger pool of candidate metrics derived from prior work and available NLP tools. We retained measures that (a) mapped cleanly onto our three research questions, (b) avoided redundancy (e.g., highly correlated variants of the same construct), and (c) behaved stably in pilot analyses on a small subset of the data. For instance, among several available readability indices, we focused on Dale-Chall and SMOG as representative measures of comprehension difficulty.

\begin{table}[ht]
\caption{Feature categories, metrics, and their theoretical meaning in response communication.}
\label{tab:metric_features}
\small
\begin{tabular}{p{3cm} p{3.2cm} p{8cm}}
\toprule
\textbf{Category} & \textbf{Metric Name} & \textbf{Measurement Explanation} \\
\midrule

\multirow[t]{6}{=}{\shortstack[l]{\textbf{Psycholinguistic}}}
& Emotional Tone & Share of affective words (positive or negative), reflecting emotional valence. \\
& Cognition \& Perception & Use of reasoning, insight, or certainty terms, indicating cognitive processing. \\
& Social Concerns & References to family, friends, or personal roles, showing social orientation. \\
& Health Concerns & Mentions of illness or health-related terms, reflecting biomedical focus. \\
& Temporal References & Use of past/present/future verbs, showing temporal framing. \\
& Informal Language & Presence of slang, netspeak, or fillers, signaling casual tone. \\
\midrule

\multirow[t]{5}{=}{\shortstack[l]{\textbf{Linguistic Structure}}}
& Verbosity & Total word and sentence counts, indicating response length. \\
& Readability & Estimated ease of comprehension using standard formulas. \\
& Complexity & Syntactic depth from dependency parsing, reflecting structural intricacy. \\
& Lexical Diversity & Vocabulary range and uniqueness, indicating linguistic variation. \\
& Repetition & Rate of repeated words, suggesting redundancy. \\
\midrule

\multirow[t]{2}{=}{\shortstack[l]{\textbf{Linguistic Style}}}
& Formality & Degree of formal grammar and tone. \\
& Analytic/Narrative Style & Dominance of analytic vs. narrative framing. \\
\midrule

\multirow[t]{4}{=}{\shortstack[l]{\textbf{Emotional Support}}}
& Politeness & LLM-derived score for courteous or softening phrases. \\
& Emotion Evaluation & LLM-estimated empathy, encouragement, and validation. \\
& Emotional Diversity & Variety of expressed emotions, measured by entropy. \\
& Emotional Composition & Distribution of emotion types (e.g., joy, sadness). \\
\midrule

\multirow[t]{5}{=}{\shortstack[l]{\textbf{Informational Support}}}
& Physician Semantic Proximity & Semantic similarity between the response and physician-authored responses, indicating informational alignment. \\
& Relative Score & Similarity to the original question, indicating relevance. \\
& Uncertainty Cues & Presence of hedging or cautious language. \\
& Information Evaluation & LLM-based score for factual or instructional content. \\
& Clinical Concept Overlap & Count of medical concepts (e.g., diseases, treatments). \\
\midrule

\shortstack[l]{\textbf{Topic Modeling}} 
& Dominant Topic & Main topic assigned by unsupervised modeling. \\
\bottomrule
\end{tabular}
\end{table}

Table~\ref{tab:metric_features} summarizes the complete set of metrics used in this study. For each feature, we specify its category, name, and measurement rationale. These metrics were introduced as part of our standard evaluation pipeline and integrated into the methodological design of \textbf{RQ1}–\textbf{RQ3}.

\subsubsection{Linguistic Feature Extraction (RQ1)}

\paragraph{Psycholinguistic}
Psycholinguistic features capture affective, cognitive, and social orientations expressed in everyday language. In caregiver-support settings, these cues help characterize \emph{how} different responders (peers, physicians, and LLM-based systems) frame their guidance and relate to caregivers, beyond the literal content of what is said \cite{tausczik2010psychological,pennebaker2015development}. For example, language indicating reasoning and certainty can shape how advice is interpreted, and affective framing (e.g., encouragement or validation) can influence how support is received \cite{hua2021emotional}. This is particularly relevant when comparing AI-generated responses with human replies in sensitive contexts such as dementia caregiving.

We operationalized these constructs using \textbf{LIWC 2022} \cite{boyd2022development}, a lexicon-based tool that maps word use to psychologically meaningful dimensions. From each response, we extracted six LIWC-based metrics: (1) \textit{Emotional Tone}, reflecting affect-laden wording; (2) \textit{Cognition \& Perception}, capturing terms related to insight, causality, and certainty (e.g., \textit{because}, \textit{think}, \textit{understand}) \cite{pennebaker2011secret}; (3) \textit{Social Concerns}, measuring references to relationships and social context (e.g., family), which are common in peer support \cite{ni2023examining}; (4) \textit{Health Concerns}, capturing illness- and clinical-related terms \cite{pavalanathan2015identity}; (5) \textit{Temporal References}, tracking past/present/future orientation as a proxy for narrative grounding and planning \cite{park2017living}; and (6) \textit{Informal Language}, including netspeak, fillers, and mild swearing, which are often associated with conversational style and approachability in online discourse \cite{squires2010enregistering}. Feature values were normalized by total word count to enable comparison across responses of different lengths.

\paragraph{Linguistic Structure}
Linguistic-structure features characterize how a response is organized and how difficult it may be to read. These properties matter in caregiving forums because caregivers often seek clear, usable guidance under stress; structure that is too sparse can reduce interpretability, while overly complex writing can increase cognitive load.

We measured five structural dimensions. \textit{Verbosity} reflects the amount of detail provided and can shape perceived thoroughness in health-related exchanges \cite{lin2017effects}. \textit{Readability} captures comprehension difficulty; simpler writing is often associated with higher engagement and trust in health contexts \cite{ermakova2014privacy}. \textit{Syntactic complexity} reflects how layered sentence constructions are, which can differ between professional and informal communication \cite{larsson2020syntactic}. \textit{Lexical diversity} measures vocabulary variation and linguistic richness \cite{richards1987type,mccarthy2005assessment}. \textit{Repetition} captures redundancy or emphasis strategies \cite{hawkins2001low}.

Operationally, \textit{Verbosity} was measured by total word count and sentence count using \texttt{textstat} (v0.7.3). \textit{Readability} was assessed with Dale-Chall and SMOG via \texttt{py-readability-metrics} (v1.2.0) and \texttt{textstat}. \textit{Syntactic complexity} was computed as average dependency parse tree depth using \texttt{spaCy} (v3.7.2; \texttt{en\_core\_web\_lg}). \textit{Lexical diversity} was quantified using Type-Token Ratio (TTR) \cite{richards1987type} and MTLD \cite{mccarthy2005assessment}, computed with \texttt{lexical-diversity} (v0.1.6). Finally, \textit{Repetition} was defined as the proportion of repeated non-stopword tokens within a response, computed from normalized counts of recurring words.

\paragraph{Linguistic Style}
Style features capture how a message is expressed (e.g., formal vs.\ conversational, narrative vs.\ analytic), which can shape perceptions of authority, relatability, and engagement in caregiver support settings. Prior work suggests that formality often signals expertise and distance, while more conversational language can support social closeness and approachability \cite{pavlick2016empirical}. Likewise, narrative language grounded in lived experience can convey empathy and shared understanding, whereas analytic language tends to emphasize structured reasoning \cite{pennebaker2014small,langum2017writing}.

We operationalized two style dimensions: \textit{Formality} and \textit{Analytic/Narrative Style}. \textit{Formality} was measured using a RoBERTa-based classifier (\texttt{s-nlp/roberta-base-formality-ranker}) implemented with HuggingFace \texttt{transformers} (v4.38.1), producing a continuous score where higher values indicate more formal responses. This aligns with evidence that LLM outputs can skew formal unless prompts explicitly elicit a more conversational tone \cite{gilardi2023chatgpt,de2021formalstyler}. \textit{Analytic/Narrative Style} was measured using the Categorical-Dynamic Index (CDI), which distinguishes categorical, structured language from dynamic, experiential discourse \cite{pennebaker2014small}. CDI was computed from LIWC 2022 outputs as:
\begin{equation}
    \text{CDI} = \text{article}_{\text{scaled}} + \text{prep}_{\text{scaled}} - \text{ppron}_{\text{scaled}} - \text{ipron}_{\text{scaled}} - \text{auxverb}_{\text{scaled}} - \text{conj}_{\text{scaled}} - \text{adverb}_{\text{scaled}} - \text{negate}_{\text{scaled}}
\end{equation}
\noindent Each term is a normalized LIWC category count scaled by total word count. Higher CDI indicates a more analytic, categorical orientation, whereas lower CDI indicates a more narrative, experiential style.

\subsubsection{Emotional and Informational Support Metrics (RQ2)}
Beyond general linguistic characteristics, we evaluated two caregiver-relevant support functions, \textit{emotional supportiveness} and \textit{informational value}, to compare how different responder groups address caregivers’ affective needs and practical questions (\textbf{RQ2}). \edit{This focus aligns with social support theory, where emotional and informational support are commonly treated as distinct (but often co-occurring) forms of support in health-related exchanges \cite{vaux1987modes,baron1990social}.} To enable consistent scoring at scale without extensive manual annotation, we used LLMs as automated judges for selected dimensions \cite{thakur2024judging}. Specifically, we used \texttt{LLaMA 3.3} for LLM-as-judge scoring, separating generator and evaluator roles to reduce shared-model bias. We then validated these LLM-based judgments with a dedicated human study (\S\ref{subsec:hil-validation}), confirming substantial alignment with expert coder ratings.

\paragraph{Emotional Support}
Emotional support includes encouragement, validation, and expressions of concern, and is a central component of caregiver support exchanges \cite{vaux1987modes}. In caregiving forums, perceived emotional quality can shape whether caregivers feel understood and whether they continue to engage with support sources \cite{jia2021online}. Because emotional support can appear both in overall stance and in finer-grained affective signals, we operationalized four complementary aspects: (1) \textit{Politeness}, (2) \textit{Emotion Evaluation}, (3) \textit{Emotional Diversity}, and (4) \textit{Emotional Composition}.

\textit{Politeness} was scored using a prompt-based classifier built with \texttt{LLaMA 3.3}. Each response was passed to the model with a schema-constrained prompt that required a structured JSON output containing: (a) a politeness score in $[0,1]$, (b) a self-rated confidence level (1-5), and (c) salient cue words supporting the score. Using the same LLaMA-based setup, \textit{Emotion Evaluation} captured whether a response explicitly conveyed empathy-adjacent behaviors (e.g., encouragement or validation). We treated these outputs as coarse-grained indicators of emotional support and evaluated their quality via human validation (\S\ref{subsec:hil-validation}).

To characterize affective content more finely, we used emotion classification under the GoEmotions taxonomy \cite{demszky2020goemotions}, \edit{grounded in Plutchik’s primary emotion theory \cite{plutchik1960multifactor}.} Using a pre-trained BERT-based model from Google, we extracted a 28-dimensional probability vector over discrete emotions (e.g., \textit{admiration}, \textit{joy}, \textit{sadness}, \textit{remorse}, \textit{confusion}). We refer to this vector as \textit{Emotional Composition}, representing the intensity profile of affective categories in each response.

We then computed \textit{Emotional Diversity} (``emodiversity'') as the entropy of the normalized GoEmotions vector $\vec{p}=(p_1,\ldots,p_{28})$. Prior work links higher emotional diversity with better emotional regulation and well-being \cite{quoidbach2014emodiversity,grossmann2019wise}. We calculated:
\begin{equation}
    \text{EmoDiv}_{28} = -\sum_{i=1}^{28} p_i \cdot \log_2(p_i)
\end{equation}
\noindent Higher values indicate a broader spread of affective categories rather than reliance on a single dominant tone. Together, these measures capture both stance-level supportiveness and the affective texture of responses across responder groups.

\paragraph{Informational Support}
Informational support refers to how much a response provides relevant, usable guidance (e.g., explanations, clarifications, or concrete suggestions) for a caregiver’s question \cite{lai2007caregivers}. In health-related settings, information that is clear and sensitive to context is especially important when caregivers are making decisions under stress \cite{mccormack2014expert}. To compare informational support across responders, we operationalized five constructs: (1) \textit{Physician Semantic Proximity}, (2) \textit{Relative Score}, (3) \textit{Uncertainty Cues}, (4) \textit{Information Evaluation}, and (5) \textit{Clinical Concept Overlap}.

\textit{Physician Semantic Proximity} assessed how closely a non-physician response (peer or AI) aligned with the physician response for the same question. We computed cosine similarity between the two responses, following the general idea of using expert alignment as a reference point when evaluating informal health content \cite{nordin2000expert}. Higher similarity indicates greater convergence in semantic space, though not necessarily better interpretability.

\textit{Relative Score} measured response relevance to the caregiver’s original question. We generated sentence embeddings using \texttt{all-MiniLM-L6-v2} (\texttt{sentence-transformers} v2.7.0) and computed cosine similarity in $[0,1]$. Higher values indicate stronger alignment with the caregiver’s stated intent, consistent with relevance-based notions of answer quality in community Q\&A \cite{jin2013users}.

\textit{Information Evaluation} was scored by prompting \texttt{LLaMA 3.3} to rate the extent to which a response provided concrete information, actionable advice, or practical suggestions. The model returned (a) a score in $[0,1]$, (b) a confidence level (1-5), and (c) a short rationale. As with emotional scoring, we validated these LLM-judge outputs against human annotations (\S\ref{subsec:hil-validation}).

\textit{Uncertainty Cues} captured whether responses signaled caution or conditionality (e.g., \textit{might}, \textit{could}, \textit{seems like}). We extracted modal/hedging markers via part-of-speech tagging in \texttt{spaCy} (v3.7.2) and counted hedging frequency per response, following pragmatic work on hedging as a strategy of epistemic humility and face-saving \cite{jingwei2013pragmatic,liu2020pragmatic}.

Finally, \textit{Clinical Concept Overlap} approximated clinical specificity by identifying biomedical entities using \texttt{scispaCy} (\texttt{en\_core\_sci\_md}, v0.5.2) \cite{neumann-etal-2019-scispacy}. We extracted concepts related to diagnoses, medications, symptoms, and procedures, and used the total number of medically relevant concepts per response as a proxy for clinically grounded language.

Together, these five measures provide a multidimensional view of informational support, spanning relevance, actionable content, uncertainty signaling, and clinical specificity, enabling systematic comparison across responder groups for \textbf{RQ2}.

\subsubsection{Topic Modeling and Contextual Analysis (RQ3)}

In addition to linguistic and support features, we examined how caregiver concerns are distributed across thematic domains using unsupervised topic modeling. This content-aware analysis addresses \textbf{RQ3}, which investigates whether the nature and quality of responses vary depending on the topics of the original caregiver question. Prior studies in online health support communities have shown that topics such as emotional burnout, medication decisions, or financial stress often elicit distinct patterns of emotional expression and informational delivery \cite{choi2020exploratory, bevan2012topic, oldenkamp2017impact}. By accounting for this variation, we aim to contextualize the linguistic and support differences observed across responder groups, including online peers, physicians, ChatGPT, and CareGPT.

We applied \texttt{BERTopic} (v0.16.1), a transformer-based topic modeling method that combines class-based TF-IDF with semantic clustering over document embeddings \cite{grootendorst2022bertopic}. This approach builds upon the foundations of Latent Dirichlet Allocation (LDA) \cite{blei2003latent}, but improves cluster quality by incorporating modern contextual embeddings. We used \texttt{DistilBERT} representations (\texttt{sentence-transformers/distilbert-base-nli-stsb-mean-tokens}) and configured the model with a minimum topic size of 15 to balance coherence and diversity. 
Each response, regardless of whether it was written by a peer, physician, or AI model, was assigned a primary topic based on the highest-probability label output by the model. For online peer caregivers, we treated each individual comment as a separate document when fitting the topic model; for physicians, ChatGPT, and CareGPT, each answer to a caregiver question was modeled as a single document. We then used BERTopic’s \texttt{visualize\_topics\_per\_class} function to compute the distribution of dominant topics across responder classes. This enabled us to identify whether different responder groups focused on distinct caregiving domains. 

\subsubsection{Human Validation of LLM-based Metric Scores}
\label{subsec:hil-validation}
To make clear that our LLM-as-judge scores are not used as unexamined proxies, we conducted a human validation study for the three judge-scored dimensions: \textit{Politeness}, \textit{Emotional Evaluation}, and \textit{Informational Evaluation}. We sampled a stratified subset of caregiver questions (covering diverse topics and intents) that remained feasible for manual coding. For each sampled question, we included the corresponding responses and asked three graduate-level coders, with prior experience analyzing online caregiver and health-support communication, to rate each dimension on a 5-point Likert scale (1=\textit{very low}, 5=\textit{very high}). We averaged ratings to form a per-item human consensus and linearly rescaled scores to $[0,1]$ to match the range of the automatic metrics.

To support consistent interpretation, coders followed a written rating rubric that defined each dimension (politeness, emotional evaluation, informational evaluation) and its anchors on the 5-point scale; they also completed a brief calibration round on held-out examples to align their use of the rubric before rating the study subset. \fredit{The rubric was developed by the research team for this study, informed by prior literature on politeness, emotional support, and informational support, and tailored to the three evaluation dimensions used in our analysis.} We assessed agreement using pairwise weighted Cohen’s $\kappa$ and pairwise Spearman rank correlation, and report these statistics to characterize reliability across coders. Finally, to evaluate alignment between human consensus and automatic scores, we computed Spearman’s $\rho$ between the human consensus and the corresponding LLM-based metric for each dimension, and inspected residuals to check for systematic bias.

\subsection{Statistical Analysis}

To examine differences in linguistic and support-related features across responder groups, we performed both statistical analyses and evidence-based interpretation. For each numeric feature relevant to \textbf{RQ1}, \textbf{RQ2}, and \textbf{RQ3}, we began with a one-way \textbf{ANOVA}, using \textit{responder group} (Online Peer, Physician, ChatGPT, CareGPT) as the independent variable. This statistical test assessed whether the population means for each group significantly differed.
When the ANOVA yielded a statistically significant result ($p <\! 0.001$), we conducted post-hoc pairwise comparisons using \textbf{Tukey’s Honestly Significant Difference (HSD)} test to identify specific group differences. Tukey’s test reports the \texttt{meandiff} between Group$_2$ and Group$_1$ (\texttt{meandiff} = Group$_2$ $-$ Group$_1$), where a positive value indicates that Group$_2$ scored higher. In contrast, we computed \textbf{Cohen’s $d$} using the reverse direction: (Group$_1$ $-$ Group$_2$) divided by the pooled standard deviation. Thus, \texttt{meandiff} and $d$ values are always of opposite sign, ensuring clear directional interpretation, e.g., a negative $d$ indicates that Group$_2$ outperformed Group$_1$.

Each ANOVA result includes the $F$-statistic, $p$-value, and the proportion of explained variance, reported as eta-squared ($\eta^2$). We interpret $\eta^2$ using conventional benchmarks \cite{cohen2016power}: 0.01 (small), 0.06 (medium), and 0.14 (large). Cohen’s $d$ effect sizes were similarly classified as 0.2 (small), 0.5 (medium), 0.8 (large), and above 1.2 (very large). We highlight findings that were both statistically significant and practically meaningful, defined as $\eta^2 > 0.14$ and/or $|d| > 1.2$.
In addition to the statistical analysis, we also conducted an interpretation of the results, focusing on the thematic richness and emotional depth of responses. After statistical testing, we inspected short representative quotes from each responder group to help interpret statistically significant differences in style, emotional support, and information quality. \edit{This targeted qualitative reading was used solely to contextualize the quantitative findings and does \textit{not} constitute a standalone qualitative analysis. }
All statistical tests were performed using \texttt{statsmodels} (v0.14.1) for ANOVA and \texttt{pingouin} (v0.5.4) for post-hoc analysis and effect size estimation. Visualization of group comparisons, including boxplots and bar plots with mean $\pm$ standard error of the mean (SEM), was performed using \texttt{matplotlib} (v3.8.2) and \texttt{seaborn} (v0.13.2).

\section{Results}
\subsection{Validation of Retrieval and Evaluation Metrics}
\label{sec:results_validation}

\subsubsection{Validation of Retrieval Performance in CareGPT}
To evaluate retrieval quality, we sampled 15 caregiver questions (15/85, $\approx$18\% of the corpus) and two annotators annotated all retrieved posts for both \textit{relevance} and \textit{influence} on the generated CareGPT response. Inter-annotator agreement was strong for both dimensions: relevance ($\kappa = 0.80$, $94.44\%$ agreement) and influence ($\kappa = 0.91$, $97.21\%$ agreement). On average, the majority of retrieved posts were judged positively, with a relevance rate of $83\%$ and an influence rate of $82\%$ (proportion of items labeled ``True'' across annotators). \edit{These findings indicate that CareGPT’s retrieved content was not only on-topic but also visibly shaped its generated responses. }
\begin{table}[hb]
\caption{Statistically significant differences in linguistic features across responder groups (RQ1), based on ANOVA (\(p < .001\), \(\eta^2 > 0.14\)) and post-hoc Tukey's test. Only group pairs with very large effect sizes (\(|d| > 1.2\)) are reported. Positive Cohen's \(d\) indicates Group\textsubscript{1} scored higher than Group\textsubscript{2}.}
\label{tab:rq1_linguistic}
\small
\begin{tabular}{llllrr}
\hline
Category & Metric & Group\textsubscript{1} & Group\textsubscript{2} & MeanDiff & Cohen's \(d\) \\ \hline

\multirow[t]{2}{*}{Psycholinguistic} 
  & \multirow[t]{2}{*}{Temporal References} 
    & \multicolumn{4}{l}{\(F = 26.16\), \(\eta^2 = 0.19\), \(p = 3.12 \times 10^{-15}\)} \\ \cline{3-6}
  &   & CareGPT   & Online Peer & 0.77     & -1.34 \\
  &   & ChatGPT   & Online Peer & 0.71     & -1.21 \\ \hline

\multirow[t]{16}{*}{Linguistic Structure} 
  & \multirow[t]{6}{*}{Verbosity (sentence count)} 
    & \multicolumn{4}{l}{\(F = 136.68\), \(\eta^2 = 0.55\), \(p = 6.87 \times 10^{-58}\)} \\ \cline{3-6}
  &   & CareGPT   & Online Peer & -8.16    & 1.63 \\
  &   & CareGPT   & Physician   & -10.96   & 2.44 \\
  &   & ChatGPT   & Online Peer & -9.15    & 1.89 \\
  &   & ChatGPT   & Physician   & -11.95   & 2.77 \\ \cline{2-6}

  & \multirow[t]{4}{*}{Verbosity (word count)} 
    & \multicolumn{4}{l}{\(F = 95.68\), \(\eta^2 = 0.46\), \(p = 8.85 \times 10^{-45}\)} \\ \cline{3-6}
  &   & CareGPT   & Online Peer & -147.51  & 1.64 \\
  &   & CareGPT   & Physician   & -170.67  & 1.84 \\
  &   & ChatGPT   & Online Peer & -156.58  & 1.82 \\
  &   & ChatGPT   & Physician   & -179.74  & 2.01 \\ \cline{2-6}

  & \multirow[t]{3}{*}{Readability (Dale-Chall)} 
    & \multicolumn{4}{l}{\(F = 103.76\), \(\eta^2 = 0.48\), \(p = 1.48 \times 10^{-47}\)} \\ \cline{3-6}
  &   & CareGPT   & Online Peer & 2.37     & -3.20 \\
  &   & ChatGPT   & Online Peer & 2.15     & -2.87 \\
  &   & Online Peer & Physician & -2.03    & 1.70 \\ \cline{2-6}

  & Complexity (Depth) 
    & \multicolumn{4}{l}{\(F = 21.05\), \(\eta^2 = 0.16\), \(p = 1.62 \times 10^{-12}\)} \\ \cline{3-6}
  &   & CareGPT   & Online Peer & -0.35    & 1.20 \\ \cline{2-6}

  & \multirow[t]{4}{*}{Lexical Diversity (TTR)} 
    & \multicolumn{4}{l}{\(F = 82.70\), \(\eta^2 = 0.43\), \(p = 4.34 \times 10^{-40}\)} \\ \cline{3-6}
  &   & CareGPT   & Online Peer & 0.06     & -1.68 \\
  &   & CareGPT   & Physician   & 0.03     & -1.56 \\
  &   & ChatGPT   & Online Peer & 0.06     & -1.70 \\
  &   & ChatGPT   & Physician   & 0.03     & -1.60 \\ \cline{2-6}

  & \multirow[t]{4}{*}{Repetition (ratio)} 
    & \multicolumn{4}{l}{\(F = 97.47\), \(\eta^2 = 0.47\), \(p = 2.10 \times 10^{-45}\)} \\ \cline{3-6}
  &   & CareGPT   & Online Peer & -0.14    & 1.85 \\
  &   & CareGPT   & Physician   & -0.15    & 1.69 \\
  &   & ChatGPT   & Online Peer & -0.15    & 2.06 \\
  &   & ChatGPT   & Physician   & -0.16    & 1.86 \\ \hline

\multirow[t]{2}{*}{Linguistic Style} 
  & \multirow[t]{2}{*}{Formality (RoBERTa)} 
    & \multicolumn{4}{l}{\(F = 44.69\), \(\eta^2 = 0.29\), \(p = 2.50 \times 10^{-24}\)} \\ \cline{3-6}
  &   & CareGPT   & Online Peer & -0.20    & 2.50 \\
  &   & ChatGPT   & Online Peer & -0.17    & 1.74 \\ \cline{2-6}

  & \multirow[t]{2}{*}{Analytic/Narrative Style (CDI)} 
    & \multicolumn{4}{l}{\(F = 46.77\), \(\eta^2 = 0.30\), \(p = 2.78 \times 10^{-25}\)} \\ \cline{3-6}
  &   & CareGPT   & Online Peer & -0.60    & 1.76 \\
  &   & ChatGPT   & Online Peer & -0.59    & 1.70 \\ \hline

\end{tabular}
\end{table}

\subsubsection{Verification of LLM-based Metric Scoring}
To assess whether LLM-as-judge metrics were qualified, we validated them against human annotations on the same 15-question subset. Human consensus ratings across politeness, emotional support, and informational support demonstrated strong inter-annotator reliability (mean $\rho = 0.81$, mean $\kappa_w = 0.78$). Comparisons with LLM-as-judge scores showed strong and statistically significant correlations: politeness ($\rho = 0.86$, $p < 0.001$), emotional support ($\rho = 0.81$, $p < 0.001$), and informational support ($\rho = 0.85$, $p < 0.001$). \edit{These alignments confirm that automated metric scores closely tracked human consensus across communicative dimensions, supporting the validity of using LLM-as-judge for large-scale evaluation.}

\subsection{Linguistic Differences Across responder groups (RQ1)}
\label{sec:results_rq1}

\begin{figure}[ht]
    \centering
    \includegraphics[width=\linewidth]{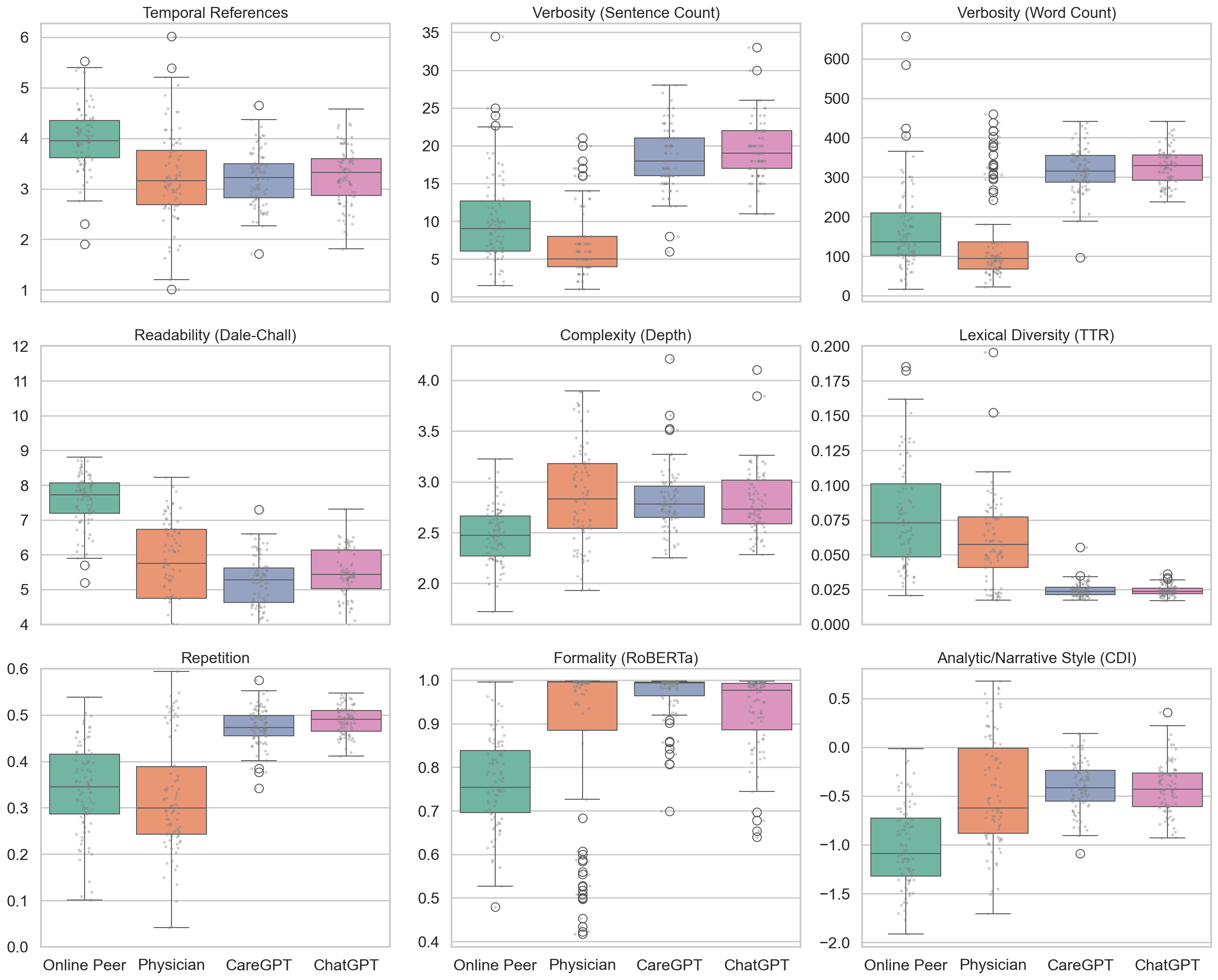}
    \caption{
    Distribution of key linguistic style features across responder groups, including verbosity, formality, lexical diversity, syntactic structure, and use of temporal references.
    }
    \Description[Boxplots of linguistic features by responder group]{
    A grid of nine boxplots comparing Online Peer, Physician, CareGPT, and ChatGPT responses across key linguistic attributes. 
    These attributes include verbosity (sentence and word count), readability (Dale-Chall), repetition, lexical diversity (TTR), categorical-dynamic index, formality (RoBERTa), temporal references, and syntactic complexity (tree depth).
    Each subplot shows the variation and central tendency of one feature across the four responder groups.
    }
    \label{fig:rq1_linguistic}
\end{figure}

We examined the linguistic characteristics of responses across four responder groups using 17 linguistic features grouped into psycholinguistic, structural, and stylistic categories. Table~\ref{tab:rq1_linguistic} summarizes statistically significant differences between groups \stat{(ANOVA $p < .001$, $\eta^2 > 0.14$)}, and we report effect sizes for pairs exhibiting large practical differences \stat{(Cohen's $|d| > 1.2$)}. Figure~\ref{fig:rq1_linguistic} visualizes the most differentiating features using boxplots. Beyond the numerical contrasts, we interpret these linguistic patterns as different ways of ``doing care'' (e.g., narrating lived experience vs.\ delivering structured guidance).

\subsubsection{Psycholinguistic Complexity}
Among psycholinguistic features, only \textbf{Temporal References} showed significant divergence. \textbf{ChatGPT} and \textbf{CareGPT} used fewer temporal cues than \textbf{Online Peers} \stat{($p=3.12\times10^{-15}$)}, with large effect sizes \stat{($d=-1.34$ for CareGPT; $d=-1.21$ for ChatGPT)}. This suggests that peer caregivers more often anchor support in time-bound trajectories (e.g., what happened \emph{recently}, what changed \emph{last week}), while LLMs more often adopt a generalized, decontextualized voice. For example, a peer may situate support with time-anchored framing such as \edit{\textit{``Last week, I visited my mom... I don’t want to be the only one...''}, while AI more often uses timeless phrasing such as \textit{``Over time, it becomes more difficult...''}.} Socially, temporal anchoring can signal ``we are in this together, right now,'' while a timeless voice can sound broadly applicable but less situated in a shared trajectory. Other psycholinguistic dimensions, such as \textbf{Emotional Tone} and \textbf{Cognition \& Perception}, were not statistically significant by our criteria \stat{($\eta^2 < 0.10$)}, and we observed no meaningful differences across responder groups. Notably, physician responses did not significantly differ from peers or AI on any psycholinguistic feature.

\subsubsection{Linguistic Structure}
Structural variation was most pronounced. \textbf{AI responses were markedly more verbose} than human replies, both in sentence count and word count. \textbf{ChatGPT} generated the longest replies, averaging nearly 12 more sentences \stat{($d=2.77$)} and 180 additional words \stat{($d=2.01$)} than physicians; \stat{these differences were highly significant ($p=6.87\times10^{-58}$)}. This verbosity can reduce the need for caregivers to ask multiple follow-ups, but it may also increase the burden of reading and triaging information, i.e., shifting labor from \textit{writing work} to \textit{interpretive work}. It also reflects a socio-technical\cite{petsolari2024sociotechnical} tradeoff in CSCW: shifting labor from humans to infrastructure \cite{cameron2021seems}. This contrast is visible in typical openings: ChatGPT often provides templated scaffolding such as \edit{\textit{``I’m really sorry you’re going through this... Here are a few steps you can consider...''}, whereas physicians more often offer compressed, workflow-oriented guidance such as \textit{``I would recommend an evaluation... Please discuss treatment options and safety planning...''}.} \edit{In this sense, AI verbosity reshapes caregiving work by reflecting AI’s lack of practical constraints on response length, while potentially increasing caregivers’ interpretive burden and the emotional labor required to process, evaluate, and act on extensive guidance under stress.}

\textbf{Readability}, assessed using the Dale-Chall readability index, also varied notably. AI-generated responses were significantly easier to read than those of \textbf{online peer caregivers}, with \textbf{CareGPT} and \textbf{ChatGPT} both about two points lower \stat{(CareGPT: $d=-3.20$; ChatGPT: $d=-2.87$)}. Because the Dale-Chall index penalizes unfamiliar vocabulary, this suggests that peer-authored responses more often use colloquialisms, shorthand, or community-specific phrasing that may feel socially resonant even when scored as ``difficult.'' In contrast, AI and physician responses tend to adopt clearer, more standardized phrasing, which may enhance accessibility at the cost of lived specificity. \edit{These differences indicate that caregivers may communicate using shared, in-group community-specific language that AI systems and clinicians may not fully recognize (e.g., ``DW/DH,'' ``POA/DPOA,'' ``MCF''), yet mismatches in vernacular across caregivers could potentially lead to misunderstandings when advice is framed in unfamiliar terms.}

\textbf{Syntactic complexity}, operationalized by average dependency parse tree depth, was modestly higher for \textbf{CareGPT} compared to \textbf{online peer caregivers} \stat{($p=1.62\times10^{-12}$; $d=1.20$)}. This indicates that CareGPT pack conditions, rationale, and recommendations into layered sentences (e.g., ``\textit{Therefore, it’s crucial to safeguard…}'' and ``\textit{If possible, consider hiring…}''), which can read as ``professional,'' but does not necessarily translate into greater clarity or engagement like online peers did (\edit{e.g., ``\textit{DH doesn’t have a diagnosis…}''}).

\textbf{Lexical diversity}, measured via the Type-Token Ratio (TTR), showed a significant drop in AI responses relative to human-written ones. Both \textbf{CareGPT} and \textbf{ChatGPT} exhibited notably lower TTR values compared to peers and physicians \stat{(all $d < -1.56$; $p=4.34 \times 10^{-40}$)}, \edit{consistent with repeated advice templates within a single reply (e.g., recurring stems such as ``\textit{consider},'' ``\textit{important},'' ``\textit{help},'' ``\textit{support},'' and ``\textit{don’t hesitate}'' across successive bullet points), whereas human responses more often vary wording through narrative detail and situational specifics.}

\textbf{Repetition}, as a distinct metric, also exhibited clear differences. \textbf{ChatGPT} was significantly more repetitive than both \textbf{online peer caregivers} \stat{($d=2.06$)} and \textbf{physicians} \stat{($d=1.86$)}, with similar trends observed in \textbf{CareGPT} \stat{(all $d > 1.69$; $p=2.10\times10^{-45}$)}. \edit{In practice, this often appears as repeated ``safety'' and ``care coordination'' framing within the same response, e.g., \textit{``...can ensure... ensure... ensure...''}.} Together, these results highlight that while LLM-generated outputs are longer and structurally polished, they tend to be less lexically diverse and more repetitive than human responses, even as they achieve greater surface readability.

\subsubsection{Linguistic Style}
Formality and discourse tone differed markedly by response type. \textbf{ChatGPT} and \textbf{CareGPT} responses were significantly more formal than those from peers \stat{(CareGPT vs.\ Peer: $d=2.50$)}, closely aligning with physicians’ tone. This suggests that despite prompt chaining or retrieval, LLMs often revert to professional discourse defaults. \edit{This alignment may also reflect the study design, as both AI systems and physicians were given comparable instructions that encouraged empathetic, consultation-style responses, whereas peer caregivers’ replies were produced organically without prompts or guidance.}

Discourse framing, measured by the Analytic/Narrative Style (CDI), revealed that peers favored narrative storytelling while AI models leaned toward analytic exposition \stat{($F=46.77$, $\eta^2=0.295$)}; \stat{CareGPT and ChatGPT differed from peers by over $d=1.69$}, emphasizing structured explanation over dynamic storytelling. This contrast is visible in how peers foreground personal trajectories, \edit{e.g., \textit{``Door locks up high work. I went through this... it was the only way I could sleep...''}, while AI foregrounds generalized guidance, e.g., \textit{``Here are a few suggestions: 1) ... 2) ... 3) ...''}. This difference likely reflects that AI systems do not have personal caregiving experience to draw on, and therefore default to analytic, generalized explanations rather than the lived, time-anchored narratives that peers use to make sense of and communicate care.}

Notably, \textbf{CareGPT showed no clear stylistic improvement over ChatGPT} in these style metrics, despite narrative prompt enhancements. Overall, RQ1 results show that each responder group enacts care through a different linguistic persona: peers signal relational closeness through narrative and time-bound lived experience, while AI and physicians emphasize legible, analytic, and formal guidance, shaping how empathy, authority, and ``work'' are distributed in caregiver support interactions. \edit{Viewed through a care infrastructure lens, these differences point to a potential division of caregiving labor in which peers contribute experiential grounding, while AI systems and clinicians provide standardized guidance that supports coordination but leaves interpretation and emotional processing to caregivers.}

\subsection{Emotional and Informational Support Patterns Across responder groups (RQ2)}
\label{sec:results_rq2}

\begin{table}[ht]
\caption{Statistically significant differences in emotional and informational support features across responder groups (RQ2), based on ANOVA (\(p < .001\), \(\eta^2 > 0.14\)) and post-hoc Tukey's test. Only group pairs with very large effect sizes (\(|d| > 1.2\)) are reported. Positive Cohen's \(d\) indicates Group\textsubscript{1} scored higher than Group\textsubscript{2}.}
\label{tab:rq2_support}
\small
\begin{tabular}{llllrr}
\hline
Category & Metric & Group\textsubscript{1} & Group\textsubscript{2} & MeanDiff & Cohen's \(d\) \\ \hline

\multirow[t]{3}{*}{Emotional Support} 
  & \multirow[t]{3}{*}{Politeness} 
    & \multicolumn{4}{l}{\(F = 76.08\), \(\eta^2 = 0.41\), \(p = 1.41 \times 10^{-37}\)} \\ \cline{3-6}
  &   & CareGPT   & Online Peer & -0.10    & 1.53 \\
  &   & ChatGPT   & Online Peer & -0.10    & 1.56 \\ \cline{2-6}

  & \multirow[t]{3}{*}{Emotion Evaluation} 
    & \multicolumn{4}{l}{\(F = 43.41\), \(\eta^2 = 0.28\), \(p = 9.81 \times 10^{-24}\)} \\ \cline{3-6}
  &   & CareGPT   & Online Peer & -0.25    & 1.35 \\
  &   & ChatGPT   & Online Peer & -0.28    & 1.66 \\ \cline{2-6}

  & \multirow[t]{3}{*}{Emotional Composition (\textit{Caring})} 
    & \multicolumn{4}{l}{\(F = 56.99\), \(\eta^2 = 0.34\), \(p = 8.31 \times 10^{-30}\)} \\ \cline{3-6}
  &   & CareGPT   & Online Peer & -0.47    & 1.74 \\
  &   & ChatGPT   & Online Peer & -0.51    & 2.06 \\ \hline

\multirow[t]{3}{*}{Informational Support} 

  & \multirow[t]{2}{*}{Physician Semantic Proximity} 
    & \multicolumn{4}{l}{\(F = 70.13\), \(\eta^2 = 0.36\), \(p = 6.11 \times 10^{-25}\)} \\ \cline{3-6}
  &   & CareGPT   & Online Peer & -0.19    & 1.52 \\
  &   & ChatGPT   & Online Peer & -0.20    & 1.62 \\ \cline{2-6}

  & \multirow[t]{2}{*}{Relative Score} 
    & \multicolumn{4}{l}{\(F = 79.87\), \(\eta^2 = 0.42\), \(p = 5.04 \times 10^{-39}\)} \\ \cline{3-6}
  &   & CareGPT   & Online Peer & -0.20    & 2.08 \\
  &   & ChatGPT   & Online Peer & -0.21    & 2.10 \\ \cline{2-6}

  & \multirow[t]{3}{*}{Information Evaluation} 
    & \multicolumn{4}{l}{\(F = 84.80\), \(\eta^2 = 0.43\), \(p = 7.22 \times 10^{-41}\)} \\ \cline{3-6}
  &   & CareGPT   & Online Peer & -0.21    & 1.51 \\
  &   & ChatGPT   & Online Peer & -0.21    & 1.51 \\
  &   & Online Peer & Physician & 0.18     & -1.23 \\ \hline

\end{tabular}
\end{table}

\begin{figure}[ht]
\centering
\includegraphics[width=0.95\textwidth]{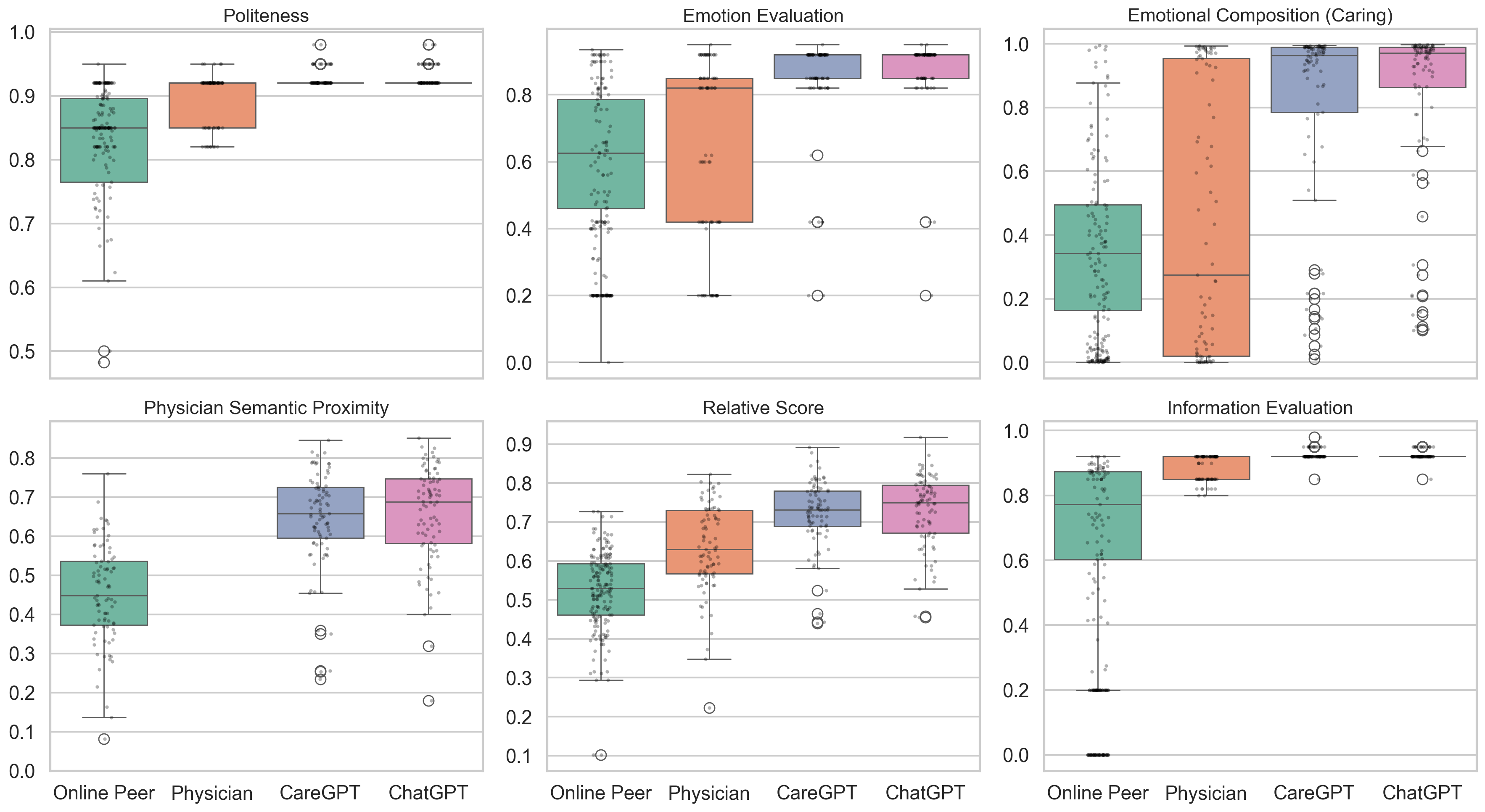}
\caption{
Boxplots illustrating key linguistic and support-related features across four responder groups: Online Peer, Physician, CareGPT, and ChatGPT. 
Top row shows \textbf{Politeness}, \textbf{Emotion Evaluation}, and \textbf{Emotional Composition (\textit{Caring})}, while the bottom row shows \textbf{Physician Semantic Proximity}, \textbf{Relative Score}, and \textbf{Information Evaluation}.}
\Description[Boxplots comparing support metrics across responder groups]{Six boxplots comparing Online Peer, Physician, CareGPT, and ChatGPT responses across key support metrics. Politeness is high for physicians and AI; emotional support is more variable among peers. 
AI shows high similarity to physician responses and questions, suggesting structured and expert-like information delivery. Peers are more diverse in style and content, indicating lived-experience variability.
}
\label{fig:rq2_features}
\end{figure}

\begin{figure}[ht]
\centering
\includegraphics[width=0.4\textwidth]{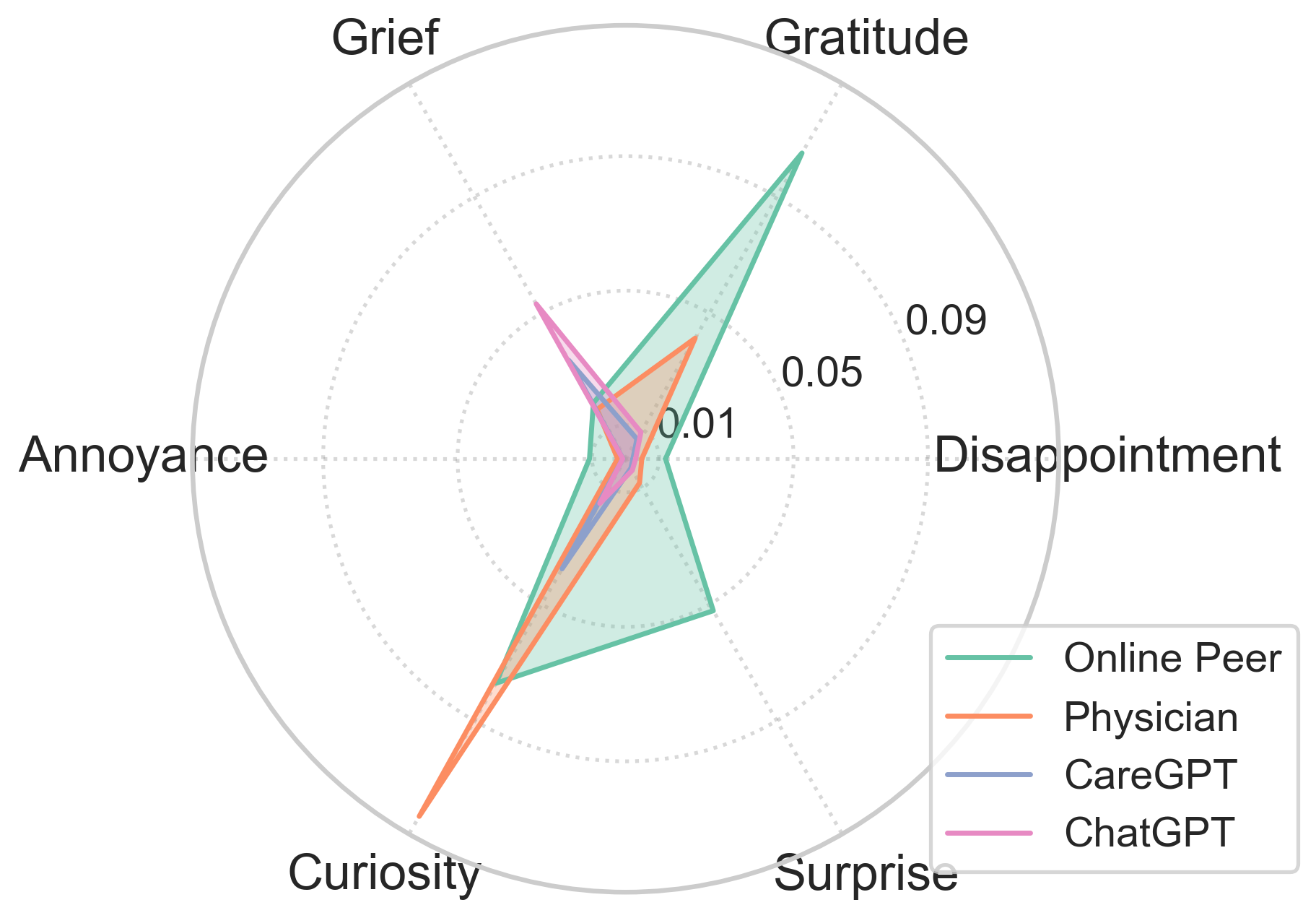}
\caption{
Radar plot showing the distribution of additional moderate-effect emotional tone dimensions across responder groups. }
\Description[Radar chart showing emotional expression differences]{
Radar chart comparing Online Peer, Physician, CareGPT, and ChatGPT emotional expression across seven categories: gratitude, grief, disappointment, surprise, curiosity, annoyance, and grief. 
Peers show the highest expression in gratitude and surprise. Physicians peak in curiosity. AI responses cluster lower overall but show slightly higher annoyance and grief than physicians.
}
\label{fig:rq2_emotionwheel}
\end{figure}

We next assess how each responder group fulfills two core support functions in this forum: providing \textbf{emotional support} and \textbf{informational support}. This analysis focuses on six features with statistically significant between-group differences, \stat{(ANOVA $p<.001$, $\eta^2>0.14$; Table~\ref{tab:rq2_support})}: three emotional support metrics (\textit{Politeness}, \textit{Emotion Evaluation}, \textit{Emotional Composition (\textit{Caring})}) and three informational support metrics (\textit{Physician Semantic Proximity}, \textit{Relative Score}, \textit{Information Evaluation}). Figure~\ref{fig:rq2_features} visualizes feature distributions across responder groups, and Figure~\ref{fig:rq2_emotionwheel} summarizes secondary emotional tones. Beyond the numerical contrasts, we interpret these patterns as different ways of enacting care and trust: standardized, risk-averse empathy and structured guidance (AI) versus relational, context-rich support grounded in lived experience (peers), with physicians often occupying a constrained but authoritative middle.

\subsubsection{Emotional Support Behaviors}

Across emotional support metrics, AI responses clustered near the top of the scale. \stat{Both \textbf{ChatGPT} and \textbf{CareGPT} scored near the ceiling on \textit{Politeness} (means $\approx0.93$), significantly higher than peers (ChatGPT vs.\ Peer $d=1.56$; CareGPT vs.\ Peer $d=1.53$; $p=1.41\times10^{-37}$).} This difference also appeared in judge confidence: \stat{LLM-based politeness confidence scores were 5.0 for ChatGPT and CareGPT, compared to 4.72 for peers and 4.98 for physicians.} In practice, AI messages repeatedly opened with templated deference and validation, \edit{e.g., \textit{ChatGPT/CareGPT: ``I’m sorry you’re going through this...''}, a reliable ``polite'' surface style that our judges consistently recognized as supportive evidence. These findings raise an open question about whether caregivers value or meaningfully benefit from AI’s highly consistent displays of politeness and empathy, particularly when such expressions are templated and detached from shared lived experience.}

A similar pattern emerged in broader \textit{Emotion Evaluation}. \stat{AI again scored higher (ChatGPT mean $\approx0.88$, CareGPT mean $\approx0.85$) than peers (mean $\approx0.60$; $p=9.81\times10^{-24}$), and AI confidence remained very high (ChatGPT 4.95; CareGPT 4.92).} However, the \emph{type} of empathy differed. AI support was often generalized and reusable, \edit{e.g., \textit{ChatGPT: ``I’m sorry to hear of the challenges you are facing... your concern is understandable.''}, whereas peer caregivers more often affirmed through situated, relational language that connects to concrete experiences, e.g., \textit{Online Peer: ``I am so sorry for your situation. I have a dear friend whose husband has FTD...''} or \textit{Online Peer: ``I am sorry. That sounds really hard...''}. Physicians, in contrast, rarely used overt emotional wording, but occasionally blended gentle validation with clinical framing, e.g., \textit{Physician: ``...cognitive decline in a loved one is stressful...''} before transitioning into next steps.}

The \textit{Emotional Composition (\textit{Caring})} metric further amplified this split. \stat{ChatGPT and CareGPT scored substantially higher than humans (Peers mean $=0.33$, Physicians mean $=0.45$, CareGPT $\approx0.81$ with $MeanDiff=0.47$ vs.\ peers, $d=1.74$; ChatGPT $MeanDiff=0.51$, $d=2.06$; $p=8.31\times10^{-30}$).} This suggests that AI responses consistently trigger ``caring'' markers according to GoEmotions, even when the message remains generic. Meanwhile, Figure~\ref{fig:rq2_emotionwheel} shows that peers (and sometimes physicians) expressed more \emph{varied} emotional tones at moderate effect sizes \stat{($p<0.001$, $\eta^2>0.06$)}, including gratitude and curiosity, interactional signals that often sustain community relationships. \edit{For instance, peers frequently acknowledged others’ efforts (\textit{Online Peer: ``thank you for this thread...''}) or asked follow-up questions to invite continued dialogue, whereas AI responses more often relied on polished sympathy statements that can drift toward clinical-sounding grief (e.g., \textit{AI: ``it’s unfortunate that...''}), potentially mismatching the poster’s context.}

Overall, RQ2 shows a distinction between \emph{performative} and \emph{relational} empathy: AI reliably produces emotionally ``acceptable'' and high-scoring support cues, while peers more often enact empathy through contextualized, sometimes messy narratives and reciprocal engagement. Physicians tend to offer limited but sincere validation constrained by a clinical, advice-forward style. Designing emotionally intelligent AI for caregiving communities thus requires more than amplifying politeness templates; it requires calibrating tone to context and supporting the kinds of relational signals that peers use to sustain trust.

\subsubsection{Informational Support Patterns}

Informational support exhibited clearer alignment between AI and physicians than emotional expression. Across \textit{Physician Semantic Proximity}, \textit{Relative Score}, and \textit{Information Evaluation}, \stat{\textbf{ChatGPT} and \textbf{CareGPT} closely tracked physician responses and exceeded peers (Table~\ref{tab:rq2_support}; Figure~\ref{fig:rq2_features}).} 

First, \textit{Physician Semantic Proximity} indicates how closely a response resembles physician-style informational content. \stat{Both AI systems scored near physicians (means $\approx0.64$-$0.67$), while peers were lower (mean $\approx0.44$; $p=6.11\times10^{-25}$; AI vs.\ Peer $d>1.5$).} This gap reflects how AI and physicians converge on standardized, referral- and resource-oriented guidance, while peers more often embed information inside narrative trajectories. For example, physician- and AI-style responses often take a referral-and-resource form (illustrative phrasing: \emph{``I would discuss this with your neurologist/PCP; they can review medications and refer you to local caregiver support and respite resources.''}), whereas peer caregivers more often fold similar information into lived timelines (illustrative phrasing: \emph{``When my husband started doing this, his doctor adjusted meds, and I also called the Alzheimer’s Association helpline—having a plan for respite made a big difference.''}).

Second, \textit{Relative Score} (response-to-question semantic alignment) highlights how tightly a responder stays anchored to the original caregiver question. \stat{ChatGPT and CareGPT achieved the highest alignment (means $=0.73$ and $0.72$), with physicians close behind (mean $=0.70$), and peers lower (mean $=0.64$; $p=5.04\times10^{-39}$; AI vs.\ Peer $d>2.0$).} In practice, AI and physicians frequently produced stepwise, on-topic answers (e.g., \textit{AI: ``Here are a few steps you can consider...''}; \textit{Physician: ``...recommend an evaluation... discuss treatment options and safety planning...''}), whereas peers often broadened the frame by sharing what worked in their own cases, which can lower semantic similarity while increasing experiential usefulness.

Third, \textit{Information Evaluation} (scored by \texttt{LLaMA 3.3}) captured perceived actionability and informativeness. \stat{CareGPT and ChatGPT scored near the upper bound (means $>0.92$), significantly higher than peers (mean $\approx0.69$; CareGPT vs.\ Peer $d=1.51$; ChatGPT vs.\ Peer $d=1.51$; $p=7.22\times10^{-41}$), while physicians also scored highly (mean $=0.93$).} Substantively, this pattern reflects how physicians and LLMs tend to present help in a standardized, stepwise format (e.g., concrete recommendations, options, and next steps), whereas peer caregivers more often prioritize contextualizing the situation—sharing what happened in their own case, expressing solidarity, and offering advice that is embedded in lived constraints rather than distilled into an ``action plan.'' Confidence ratings reinforced this difference in stance: \stat{ChatGPT, CareGPT, and physicians all had mean confidence $=5.0$ (std $=0$), while peer confidence was lower and more variable (mean $=4.52$, std $=0.78$),} consistent with peers positioning their guidance as contingent and experience-based rather than authoritative.

These aggregate differences align with how each group ``packages'' information. AI responses frequently recommended named institutional resources and standardized care pathways (e.g., \edit{\textit{AI: ``...contact the Alzheimer’s Association...''} or \textit{AI: ``...reach out to your primary care doctor or neurologist...''}), reflecting a formal guidance persona. CareGPT also foregrounded caregiving logistics common in forum data (e.g., \textit{CareGPT: ``...adult day care or respite care...''}). Physicians similarly emphasized diagnostics, medication, referrals, and coordination, often concisely and with clear authority. Peers contributed community-specific practical knowledge, frequently referencing legal/coordination realities such as POA and care placement, and embedded such information in lived trajectories (e.g., \textit{Online Peer: ``...Do you have financial POA?...''} or stories about navigating symptoms and family dynamics). Even when these narratives drift from the immediate question (lowering semantic alignment), they can surface constraints and ``what it actually looks like'' details that structured responses omit.}

Overall, RQ2 results \edit{suggest a division of labor in informational support}: AI and physicians deliver consistently organized, high-density guidance that supports cognitive trust (clarity, coverage, actionability), while peers contribute contextualized, community-grounded knowledge that often supports relational trust (shared reality, local constraints, ``what worked for me''). Rather than treating ``more AI empathy'' as the primary design goal, these findings motivate hybrid designs that orchestrate strengths: structured guidance and resources from AI/clinicians paired with peer-sourced experiential nuance that helps caregivers interpret and act on that guidance in real life.

\subsection{Topic-Specific Performance of AI vs. Human Responders (RQ3)}
\subsubsection{Topic Modeling}

\begin{figure}[ht]
\centering
\includegraphics[width=0.95\textwidth]{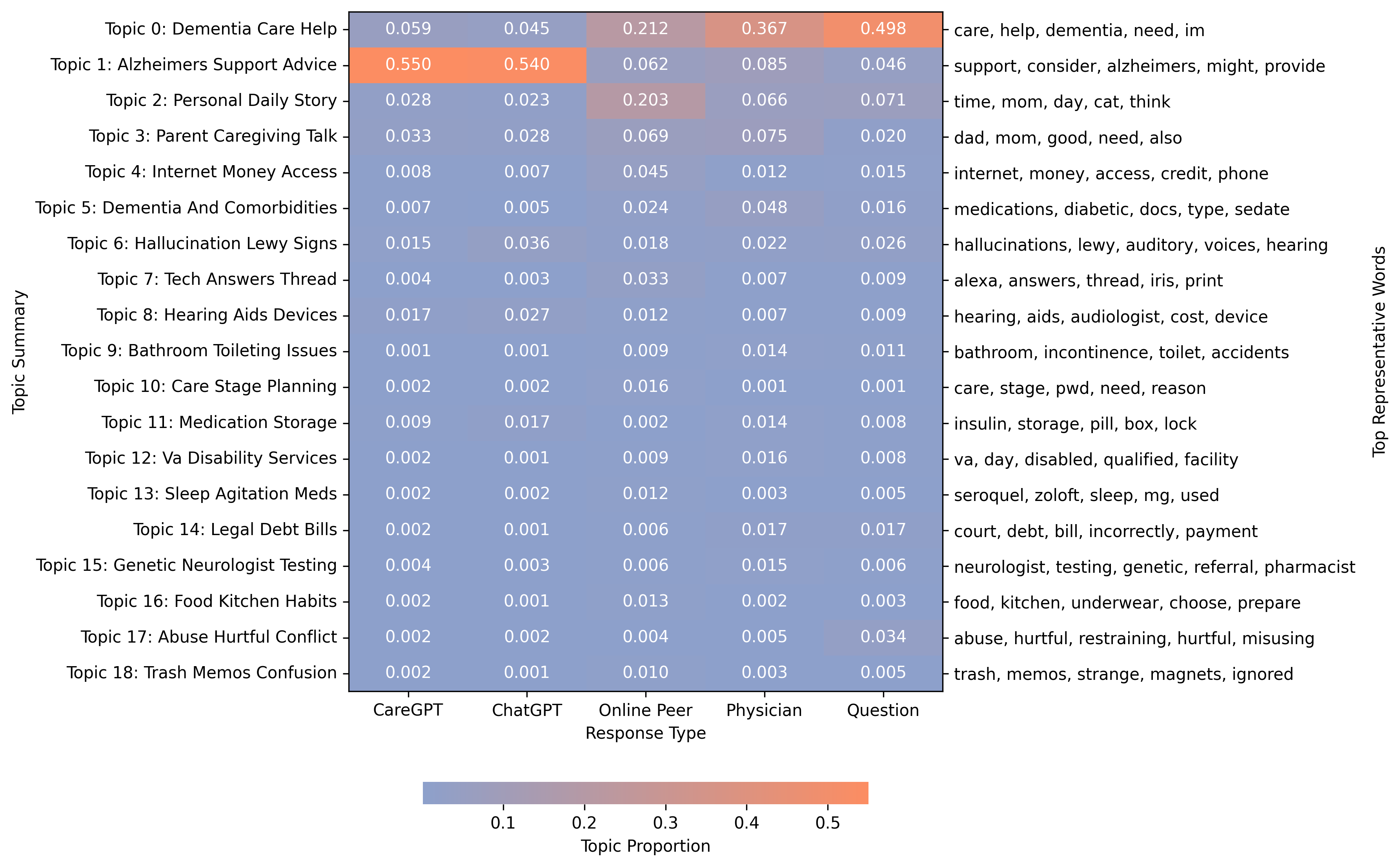}
\caption{
Average topic proportions across responder groups. Each cell presents the mean topic probability assigned to responses of each type, based on BERTopic outputs. Topics are derived from clustering response content across the full dataset, labeled with brief summaries and representative keywords.
}
\Description[Heatmap of topic proportions by response type]{
Heatmap showing average topic probabilities across four responder groups, plus original questions. Y-axis lists 19 topics. Color intensity and overlaid numeric values show how frequently each response type addresses each topic.}
\label{fig:topic_distribution_heatmap}
\end{figure}

To explore the topics of responses across different responder groups, we applied the BERTopic algorithm to the entire corpus of replies and questions. This unsupervised topic modeling approach yielded 19 coherent topics, determined by BERTopic’s hierarchical clustering algorithm, with each characterized by a distribution of representative words and most representative documents. Figure~\ref{fig:topic_distribution_heatmap} presents a heatmap of the average topic probability per response type. These values represent the average share of each topic within all responses of a given type, providing a high-level view of content alignment with responder roles.

Several patterns are evident from the heatmap. First, Topic 0 (``\textit{Dementia Care Help}'') dominates the question distribution ($mean = 0.498$), consistent with the forum’s overall focus on practical caregiving concerns. Physicians also show the highest average proportion for Topic 0 ($mean = 0.367$), reflecting their alignment with general care inquiries. In contrast, both CareGPT and ChatGPT concentrate on Topic 1 (``\textit{Alzheimer's Support Advice}''), a domain focused on direct, prescriptive recommendations. Their strong emphasis on this topic (mean $>$ 0.54) suggests a tendency to interpret questions as advice-seeking prompts and to respond with structured, guidance-oriented content. Online Peer responses, on the other hand, are less centered on these general or directive themes and show markedly higher engagement with Topic 2 (``\textit{Personal Daily Story}'', $mean = 0.203$), and also Topic 3 (``\textit{Parent Caregiving Talk}''), and Topic 4 (``\textit{Internet Money Access}''). These patterns point to a peer preference for narrative sharing and personal situational reflections rather than clinical or instructional replies. These findings align with our RQ2 results, which showed that peer caregivers tend to provide emotionally nuanced, community-based support, whereas AI systems emphasize structured politeness and high-confidence factual suggestions. The topic-level distributions further reinforce this divide in support styles.

\subsubsection{Topic Proportions}

\begin{figure}[hb]
\centering
\includegraphics[width=0.9\textwidth]{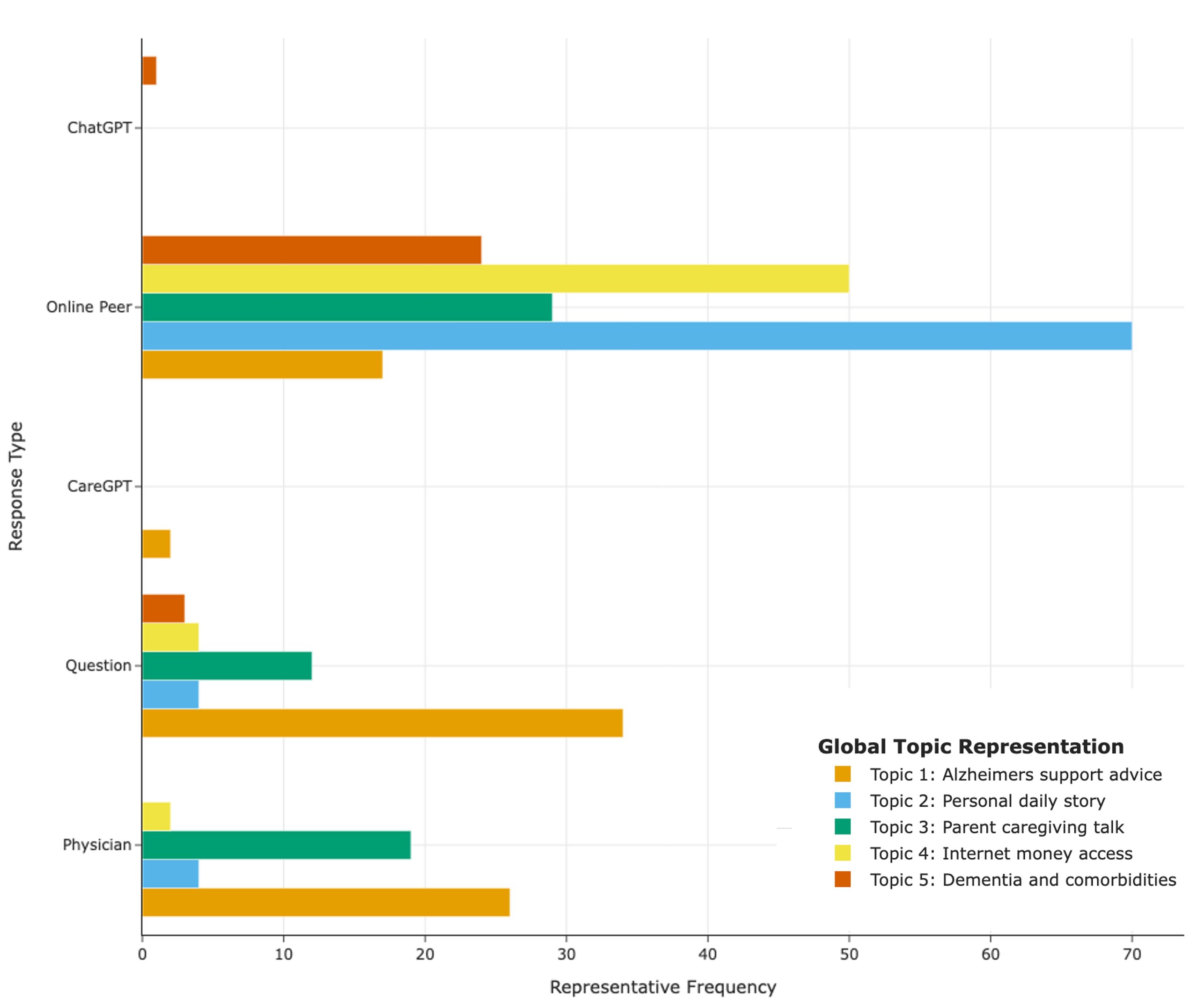}
\caption{
Distribution of the top five most representative topics across responder groups, based on BERTopic’s \texttt{topics\_per\_class} function. 
Each bar represents the frequency of responses primarily associated with each of the selected topics (Topics 1–5) for a given responder group.
}
\Description[Bar chart of top 5 topics by response type]{
This bar chart compares how frequently five major topics, Alzheimer's support advice, personal daily story, parent caregiving talk, internet money access, and Dementia and Comorbidities, are represented in responses by each group: ChatGPT, Online Peer, CareGPT, Question, and Physician. Each colored bar corresponds to a topic, and the x-axis represents the representative frequency. Online peers are shown with the highest counts across several topics.
}
\label{fig:topics_per_class_bar}
\end{figure}

To move beyond average topic probabilities and capture which topics most define each response, we leveraged BERTopic’s \texttt{topics\_per\_class} function. This approach identifies the most representative topic for each response document and counts their distribution across responder groups. The resulting bar chart (Figure~\ref{fig:topics_per_class_bar}) shows how many responses of each type were primarily associated with each of the top 5 topics (excluding the generic Topic 0).
The plot illustrates striking differences in topical scope and specialization across responder groups. \textbf{Online Peer responses demonstrate the greatest topical breadth}, with substantial representation across all five shown topics. Notably, over 70 responses were best matched with Topic 2 (``\textit{Personal Daily Story}''), highlighting the peer community's emphasis on lived experience and empathetic sharing. Substantial numbers of peer responses also centered on Topic 3 (``\textit{Parent Caregiving Talk}''), Topic 4 (``\textit{Internet Money Access}''), and Topic 5 (``\textit{Dementia and Comorbidities}''), suggesting a diversity of concerns addressed informally and practically. 
By contrast, CareGPT responses were almost entirely aligned with Topic 1 (``\textit{Alzheimer's Support Advice}''), indicating a narrow but consistent pattern of structured caregiving guidance. ChatGPT responses were limited to Topic 5, focusing on treatment-specific content, such as medication or symptom management. No other topics appeared as the most frequent or defining themes in either LLM’s output, suggesting that current AI systems may follow strongly scripted or training-biased topic patterns rather than adapting to a variety of caregiver concerns.

Physician responses were more topically balanced, appearing across Topics 1–4, but notably absent from Topic 5. This gap may be meaningful; physicians may hesitate to discuss chronic illness comorbidities like diabetes in dementia-focused forums, sticking instead to domain-relevant or directly observable issues such as caregiving advice and financial burden.
Lastly, questions posted in the forum showed limited topic diversity, but interestingly had primary alignment  with Topic 1 (caregiving advice), as well as Topics 2 (personal story), 4 (finance-related issues), and 5 (treatment queries). This confirms that the original questions span both emotional and practical needs, and reaffirms the need for diverse responder groups to cover this range.
Overall, these findings show that LLM-generated responses are topically narrow but predictable, while peer responses are wide-ranging and adaptive to many caregiver concerns. Physicians occupy a middle ground, staying largely within their domain of expertise.

\section{Discussion}
This work provides a systematic, side-by-side comparison of peer, clinician, and AI responses to the same ADRD caregiver questions. \edit{We contribute to CSCW discussions of AI participation in caregiving communities by characterizing \fredit{role-differentiated care work} across a hybrid care infrastructure: peers, clinicians, and AI each enact support through distinct linguistic forms, trust cues, and topical behaviors.} Guided by prior work on care infrastructures, emotional and informational labor, and socio-technical trust calibration, we synthesize (i) overarching empirical patterns, (ii) mechanisms that plausibly produce these patterns, and (iii) implications for designing hybrid support arrangements where AI augments rather than replaces human care work.

\subsection{\edit{Role Differences Across Peers, Clinicians, and AI}}

\begin{table}[ht]
\caption{Empirically observed strengths and limitations by responder group, illustrating how support work is distributed across a hybrid care infrastructure.}
\label{tab:distributed-care-summary}
\small
\begin{tabular}{p{2.2cm} p{3.75cm} p{3.75cm} p{3.75cm}}
\hline
\textbf{Responder group} & \textbf{Strengths (observed)} & \textbf{Limitations (observed)} & \textbf{Role in distributed care} \\
\hline

\edit{\textbf{Peer caregivers}} &
\edit{Time-anchored narratives; community vernacular; broader topic range (e.g., emotional, daily logistics, legal and financial)} &
\edit{Less standardized structure and readability; lower physician-style proximity and information-evaluation} &
\edit{Lived context and relational legitimacy for sensemaking} \\
\hline

\edit{\textbf{Physicians}} &
\edit{Concise, clinically bounded guidance; high physician-style informational content} &
\edit{Less narrative and relational work; emotional support often brief and secondary} &
\edit{Accountable clinical guidance and care pathways} \\
\hline

\edit{\textbf{ChatGPT}} &
\edit{Highly structured and readable; high information-evaluation and physician proximity; consistently high politeness} &
\edit{Low temporal and narrative grounding; repetitive and templated phrasing; advice-heavy topical focus} &
\edit{On-demand “consultation-style” organization of options and resources} \\
\hline

\edit{\textbf{CareGPT}} &
\edit{Tracks ChatGPT on informational and emotional metrics; retrieval-conditioned on peer discussions} &
\edit{Minimal measurable separation from ChatGPT; continued formal, advice-forward stance} &
\edit{Reinforces LLM role as structured guidance (not peer-like lived experience)} \\
\hline

\end{tabular}
\end{table}

\edit{We read these results through distributed care infrastructure: differences in linguistic style, emotional expression, and informational framing point to how caregiving labor is allocated and coordinated across people and technologies, not simply to “better” or “worse” writing. Table~\ref{tab:distributed-care-summary} summarizes the patterns by making the tradeoffs explicit across responder groups, consistent with CSCW accounts of care infrastructures as relational and contingent systems that rely on articulation, sensemaking, and ongoing mutual adjustment rather than uniform service provision \cite{karusala2023unsettling}. Within this framing, AI participation is characterized by standardized, polite, and scalable support that is less tied to lived trajectories and community language. This consistency can lower interactional barriers and make resources easier to access, but it also shifts more interpretive and emotional work onto caregivers, echoing CSCW descriptions of sociotechnical tradeoffs where infrastructure absorbs some labor while generating new demands elsewhere \cite{cameron2021seems}. The coexistence of peer narratives, clinically bounded physician guidance, and AI-mediated support also aligns with social support theory, which treats emotional and informational support as multiple, co-present forms rather than a single support function \cite{scott2014social}. These findings therefore caution against treating AI as a standalone solution in affective or high-stakes contexts \cite{liu2022will, liu2025regulation}, and motivate hybrid caregiving systems that support coordination across distributed actors and make the limits of AI-mediated empathy visible within the care infrastructure.} \fredit{These patterns indicate that care support is organized through complementary forms of relational and informational labor, rather than through convergence toward a single, idealized style of response.} In this sense, \fredit{linguistic differences matter not only as descriptive features of responses, but as cues through which caregivers may interpret relational stance, legitimacy, and the fit of a response to a given support need.}

Across RQ1-RQ3, we observe a consistent division of communicative labor in how care work is enacted through language. First, \textbf{peers} disproportionately contribute \emph{narrative, temporally anchored lived-experience support}. They often situate advice in personal trajectories (\edit{e.g., ``my mom passed away in 2020,'' ``I took early retirement to care 24/7,'' ``some days are so much worse than others''}), using time markers and first-person storytelling to signal solidarity and shared reality. This aligns with CSCW observations that online health communities value lived experience and being heard as core signals of support and belonging \cite{bates2023empathy,massimi2014life,smriti2024emotion}, and that such narrative participation can sustain trust and legitimacy within peer infrastructures \cite{ni2025catalysts}. Second, \textbf{clinicians} tend to provide \emph{concise, advice-forward, clinically bounded guidance}, reflecting professional norms of authority and clarity \cite{ding2020getting}. Their support often takes the form of referral and scoped actionability (e.g., ``the Alzheimer's Association has wonderful resources''; ``an evaluation of decision-making capacity... will also help for legal decision making''). This style can increase perceived credibility, but it may offer less reciprocal vulnerability and less colloquial warmth than peers, reflecting tensions documented in CSCW health communities between expertise and affirmation \cite{smriti2024emotion,brewer2021challenging,poon2021computer}. Third, \textbf{LLMs (ChatGPT and CareGPT)} consistently enact a \emph{formal, verbose, structured guidance persona}. In grief-support scenarios, the models often open with acknowledgement \edit{(e.g., ``I'm very sorry for your loss''), then enumerate institutional pathways (e.g., ``a 24/7 helpline,'' ``local chapters,'' ``bereavement services''), and close with generalized reassurance (e.g., ``you are not alone'')}. This echoes findings that LLMs default to essay-like, highly structured support styles \cite{shahid2024examining, yin2024should}. In our setting, this yields consistently readable and “acceptable” surface cues, yet it can also read as templated or socially misaligned with community norms.

These three voices represent complementary but non-substitutable forms of care work: peers offer relational grounding through time-anchored narratives; clinicians offer bounded authority and clinically accountable pathways; AI offers scalable structure, coverage, and consistently on-topic scaffolding. \fredit{The theoretical insight is therefore not merely that these responder groups differ, but that they enact distinct forms of caregiving labor and legitimacy within the same support ecology.} Importantly, our results suggest that AI's strongest observable contribution is not ``becoming human-like,'' but producing consistent, organized, action-oriented support that reduces search and synthesis burden, while human voices (peer and clinician) supply situated judgment, accountability, and community legitimacy.

\textbf{Design Implications.} \fredit{These findings motivate design considerations for systems that account for the complementary roles of peers, clinicians, and AI, rather than treating AI as a replacement for human care work.} Systems should maintain the legitimacy of peer narratives (including their willingness to surface non-clinical constraints such as POA, guardianship, or insurance risks that arise in ``unstable'' family environments), protect clinicians' bounded authority and uncertainty signaling, and position AI as a transparent infrastructure layer that provides scaffolding (e.g., resource routing, checklists, summarization) without competing with humans for identity or interpersonal status \cite{liu2025regulation}. Rather than optimizing for a single ``best'' response, CSCW designs can specify which participant is best suited to provide which type of support, and how these contributions are presented to caregivers.

\subsection{\edit{Sources of Differences in Language, Trust Cues, and Topic Scope}}

We interpret these patterns through three related reasons that shape how each actor ``does care''.

\edit{\textbf{(1) Norms and constraints produce distinct authenticity signals.}
Peers operate within a community norm where credibility is often earned through lived experience, temporal anchoring, and reciprocal disclosure \cite{bates2023empathy,massimi2014life,smriti2024emotion}.} In our corpus, peers frequently mark legitimacy by narrating concrete trajectories and recent incidents \edit{(e.g., ``over the last few years it has been getting worse,'' ``last week I came home and her car was left on overnight'')}, positioning advice as emerging from ongoing, situated responsibility rather than abstract expertise. Clinicians operate under professional constraints that prioritize clinical accuracy and appropriate scope \cite{ding2020getting}, which can limit narrative participation even when emotional support is needed \cite{smriti2024emotion,brewer2021challenging,poon2021computer}. Their supportive language often appears as bounded empathy plus action-oriented triage (e.g., acknowledging that dementia diagnoses are hard, then shifting to self-care, primary-care follow-up, therapy, medication options, and crisis escalation when needed). In contrast, LLMs can generate empathy at scale, but that empathy often appears as stable templates and polished phrasing \cite{shahid2024examining, yin2024should}. This makes AI ``reliably polite'' yet potentially vulnerable to perceived inauthenticity, a form of artificial empathy that can feel ungrounded in shared experience \cite{seitz2024artificial}.

\edit{\textbf{(2) Trust has more than one dimension, and peers, clinicians, and AI tend to rely on different cues.}}
Our findings are consistent with the distinction between emotional trust (``they understand me'') and cognitive/epistemic trust (``they know what they are talking about'') \cite{kyung2022rationally}. Peers often earn emotional trust through contextualized empathy and pragmatic, experience-linked warnings (e.g., noticing safety-relevant slips like leaving a car running, or anticipating family disagreement about symptom severity). Clinicians and AI more strongly signal cognitive trust via structured, advice-forward content and expert-like framing, including referrals, formal coping strategies, and clear boundary conditions for urgent escalation. However, pushing AI too far toward human-like affect can trigger discomfort akin to an ``uncanny valley'' of empathy \cite{misselhorn2009empathy}, while under-expressive AI may be dismissed as cold. \edit{This tension motivates explicit trust-calibration features (e.g., transparency, sourcing, and appropriate disclaimers) that help caregivers interpret AI behavior \cite{shen2024empathy,huang2024fact} and develop strategies under uncertainty \cite{lopez2025users}.}

\edit{\textbf{(3) Topic scope reflects how each responder frames what the question is asking for.}}
Peers treat caregiving as a holistic life condition; their topical breadth reflects the way caregiving problems spill across emotional, social, legal/financial, and medical domains \cite{nakikj2017park}. This also defines ``expertise boundary'': peers may explicitly caution against informal legal advice and redirect caregivers toward certified elder-law resources (e.g., emphasizing that guardianship is costly/contested and recommending CELA/NAELA directories) while still sharing forum-grounded heuristics. Clinicians remain more bounded by domain remit \cite{ding2020getting}. AI, especially under an office-framed instruction, tends to interpret posts as requests for prescriptive guidance, reinforcing a narrower, consultation-like posture \cite{shahid2024examining, yin2024should}. In other words, topic breadth is not only ``what knowledge one has,'' but \emph{how one frames the interactional purpose} of the post.

\fredit{These reasons indicate that differences across peers, clinicians, and AI are not merely stylistic. Rather, they reflect how different responder groups enact support roles under different norms, constraints, and expectations for what counts as credible or authentic care.} This is why simply making AI ``more empathetic'' is unlikely to reproduce the relational credibility of peer infrastructures \cite{seitz2024artificial}, and why hybrid designs must support ongoing trust calibration rather than assuming stable trust \cite{lopez2025users}.

\subsection{Design Implications and Interpreting CareGPT’s Similarity to ChatGPT}

\fredit{Our results suggest a practical design direction: support systems should be designed not only to combine what peers, clinicians, and AI each do well, but also to preserve their distinct roles rather than treat them as interchangeable.} Peers, clinicians, and AI contribute different kinds of value. Peers bring lived constraints and relational resonance. Clinicians bring bounded authority and accountability. AI provides scalable structure and broad informational support \cite{kyung2022rationally}. The design question is how to present these voices together in a way that preserves their distinct roles.

This also matters for trust. Interfaces need to help caregivers interpret support, rather than assuming trust is something a system can simply \emph{earn}. Caregivers develop strategies and folk theories for judging advice under uncertainty \cite{lopez2025users}. Hybrid systems should therefore make AI status and limits clear through trust-calibration features such as transparency about data resources, links to authoritative references, and simple disclaimers that distinguish general guidance from clinician-appropriate decisions \cite{shen2024empathy,huang2024fact}. At the same time, systems should avoid over-claiming emotional competence. Overly human-like empathy can feel uneasy or inauthentic when it is not grounded in shared experience \cite{seitz2024artificial,misselhorn2009empathy}. These supports help caregivers decide when to use AI as a first pass, when to defer to a clinician, or when to seek peer input that accounts for family dynamics and practical constraints. \fredit{Prior work also shows that users often choose AI precisely because of its non-human traits \cite{park2025current}. This aligns with our finding that AI's strongest contribution lies not in mimicking human empathy but in offering scalable availability and structured guidance that complements rather than replaces human care work.}

This complementarity is visible when the same caregiver question receives different kinds of help. Clinicians tend to respond with concise, clinically bounded guidance, often pointing to evaluations, referrals, or safety steps, consistent with professional expectations around scope and responsibility \cite{ding2020getting}. Peer caregivers more often respond by describing what similar situations look like in their own lives over time, including how grief, exhaustion, and family tension accumulate alongside daily tasks. These experience-anchored narratives do more than convey information. They signal recognition, legitimacy, and solidarity in ways that support communities consistently value \cite{bates2023empathy,massimi2014life,smriti2024emotion}. AI responses follow a different pattern. They typically organize support into structured, step-by-step suggestions paired with polite reassurance and resource pointers, producing replies that are easy to follow and consistently read as empathetic \cite{shahid2024examining, yin2024should}. These differences do not point to a single ideal response style, but \fredit{show} how caregiving support is built through multiple forms of help: AI surfaces options and common pathways, clinicians set medical boundaries and accountability, and peers translate guidance into lived realities while sustaining participation in the community \cite{nakikj2017park}.

\textbf{Design Implications.} First, in caregiving contexts, RAG should be treated as a grounding component rather than a guarantee of peer-like relational support. If the goal is to better align with community norms, the system may require instructions tailored to specific domains, example-based response shaping, or clearer role definitions to encourage it to consider specific contexts (e.g., time, relationships, and constraints) rather than providing generic advice \cite{shahid2024examining, yin2024should}. Second, a promising path is to use AI in a support role for human responders \fredit{and to optimize it for what it does best: structured, instrumental support}. For example, AI can draft structure and resource-oriented content, while peers and clinicians add lived specificity, \fredit{authentic empathy,} boundary-setting (e.g., when certified legal or clinical advice is required), and accountability. This preserves transparent non-human status while augmenting, rather than displacing, the emotional and informational labor embedded in caregiving communities \cite{liu2025regulation}.

CareGPT’s similarity to ChatGPT is itself a design result. Across RQ1-RQ3, retrieval-plus prompt-injection alone did not measurably shift the model’s communicative stance. Even when community discourse was available, the system often retained the same consultation-like, polished guidance role. This indicates that adding retrieved peer text is not sufficient to align model behavior with community practice. If the goal is to change how the model frames its role, stronger signals are needed, such as role framing and interactional expectations that shape what counts as an appropriate response \cite{sharma2023human}. RAG may improve topical grounding, but it does not by itself produce peer-like narrative positioning or relational credibility.

\subsection{Limitations and Future Work}
This study has several limitations that motivate future research. First, our data are drawn from a single online community; caregiving norms vary across platforms, cultures, and languages. Replication across other caregiving and health communities is needed. Second, our AI responses reflect models and configurations available at the time of study; rapid model evolution may change observed patterns, reinforcing the need for ongoing evaluation. \edit{Third, although we validated LLM-as-judge metrics against human ratings, automated judgments cannot fully capture caregivers’ subjective experience of being supported. Future work should incorporate caregiver-facing evaluations (e.g., preference testing, perceived emotional safety, and trust calibration effects) and examine how interface scaffolds influence uptake.} Finally, we focus on message-level comparisons rather than longitudinal trajectories. Future longitudinal and ethnographic work could theorize how hybrid care infrastructures evolve over time: when AI stabilizes practices, when it introduces new burdens or dependencies, and how communities negotiate governance and accountability as non-human actors perform visible care work.

\section{Conclusion}
In this study, we compared how \textit{peer caregivers, physicians, ChatGPT}, and \textit{CareGPT} responded to the same set of Alzheimer's disease caregiver questions, offering the first controlled benchmark of human and AI voices in this sensitive context. This allows us to move beyond hypothetical debates to concrete evidence of what AI currently does well and where it falls short. We found that peer caregivers contribute empathy through lived experiences, physicians provide trusted accuracy, and AI offers content consistency. While AI models can emulate human knowledge and polite tone, they lack the personal nuance and contextual adaptability of peers and physicians.  \fredit{These findings contribute to CSCW theory by showing that peer caregivers, physicians, and AI systems occupy distinct but complementary roles within caregiving support infrastructures, and that these roles reflect different forms of relational, informational, and accountable support labor. The analytical pipeline is also transferable to other domains in future work.} Overall, our findings suggest that human-AI collaboration in caregiving is best seen as integration rather than replacement.

These findings point to sociotechnical systems that strengthen, rather than undermine, the values of caregiver communities and domain experts by using AI \fredit{where it can provide support credibly}, while preserving space for \fredit{peer empathy and clinician accountability}.

\begin{acks}
Sean S. Huang and James S. Powers were partly supported by the HRSA Geriatric Workforce Enhancement Program, Award No. 6 U1QHP33085-05-07.

Portions of this manuscript were revised for clarity and presentation with assistance from OpenAI's ChatGPT (GPT-5.2; accessed January 2026). The tool was used only to rephrase and reorganize author-written text based on author-provided ideas and previous drafts. All analyses, interpretations, and conclusions remain those of the authors, who take full responsibility for the accuracy and integrity of the content.
\end{acks}

%%%%%
\bibliographystyle{ACM-Reference-Format}
\bibliography{ref}

\end{document}